\documentclass[10pt,twocolumn,letterpaper]{article}

\usepackage{cvpr}
\usepackage{times}
\usepackage{epsfig}
\usepackage{graphicx}
\usepackage{amsmath}
\usepackage{amssymb}
\usepackage{amsfonts}
\usepackage{bm}
\usepackage[dvipsnames]{xcolor}
\usepackage{booktabs}
\usepackage{subfig}
\usepackage{xfrac}
\usepackage{threeparttable}
\usepackage{multirow}
\usepackage{rotating}
\usepackage{changepage}
\usepackage{makecell}
\usepackage{floatrow}
\newcolumntype{P}[1]{>{\centering\arraybackslash}p{#1}}
\usepackage{lineno}
\modulolinenumbers[5]

\usepackage[breaklinks=true,bookmarks=false]{hyperref}

\cvprfinalcopy 

\def\cvprPaperID{6246} 

\begin{document}

\title{A Relation-Augmented Fully Convolutional Network for Semantic Segmentation in Aerial Scenes}

\author{Lichao Mou$^{1, 2*}$, \quad Yuansheng Hua$^{1, 2}$\thanks{Equal contribution}, \quad Xiao Xiang Zhu$^{1, 2}$\\
$^{1}$ Remote Sensing Technology Institute (IMF), German Aerospace Center (DLR), Germany\\
$^{2}$ Signal Processing in Earth Observation (SiPEO), Technical University of Munich (TUM), Germany\\
{\tt\small \{lichao.mou, yuansheng.hua, xiaoxiang.zhu\}@dlr.de}
}
\maketitle

\begin{abstract}
   Most current semantic segmentation approaches fall back on deep convolutional neural networks (CNNs). However, their use of convolution operations with local receptive fields causes failures in modeling contextual spatial relations. Prior works have sought to address this issue by using graphical models or spatial propagation modules in networks. But such models often fail to capture long-range spatial relationships between entities, which leads to spatially fragmented predictions. Moreover, recent works have demonstrated that channel-wise information also acts a pivotal part in CNNs. In this work, we introduce two simple yet effective network units, the spatial relation module and the channel relation module, to learn and reason about global relationships between any two spatial positions or feature maps, and then produce relation-augmented feature representations. The spatial and channel relation modules are general and extensible, and can be used in a plug-and-play fashion with the existing fully convolutional network (FCN) framework. We evaluate relation module-equipped networks on semantic segmentation tasks using two aerial image datasets, which fundamentally depend on long-range spatial relational reasoning. The networks achieve very competitive results, bringing signiﬁcant improvements over baselines.
\end{abstract}

\section{Introduction}
\begin{figure}[t]
    \centering
    \includegraphics[width=1\textwidth]{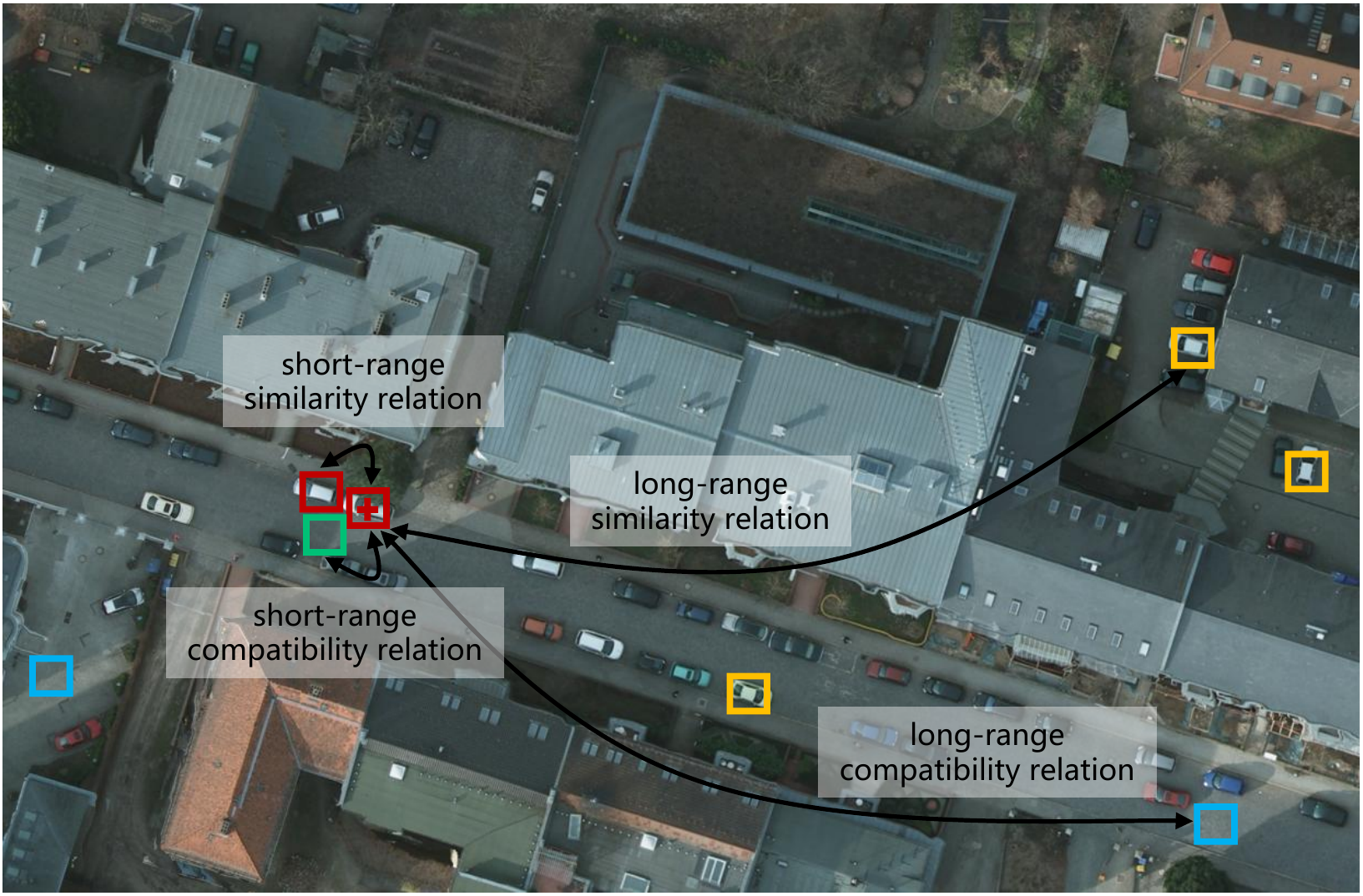}
    \caption{Illustration of long-range spatial relations in an aerial image. Appearance similarity or semantic compatibility between patches within a local region (red--red and red--green) and patches in remote regions (red--yellow and red--blue) underlines our global relation modeling.}
    \label{fig:fig1}
\end{figure}
\label{sec:intro}
Semantic segmentation of an image involves a problem of inferring every pixel in the image with the semantic category of the object to which it belongs. The emergence of deep convolutional neural networks (CNNs) \cite{krizhevsky2012imagenet, simonyan2014very, he2016deep, huang2017densely, 7803544, cas_l4} and massive amounts of labeled data has brought significant progress in this direction. However, although with more complicated and deeper networks and more labeled samples, there is a technical hurdle in the application of CNNs to semantic image segmentation---contextual information.
\par
It has been well recognized in the computer vision community for years that contextual information, or \emph{relation}, is capable of offering important cues for semantic segmentation tasks~\cite{Gould, fidler}. For instance, spatial relations can be considered semantic similarity relationships among regions in an image. In addition, spatial relations also involve compatibility and incompatibility relationships, \ie, a vehicle is likely to be driven or parked on pavements, and a piece of lawn is unlikely to appear on the roof of a building. Unfortunately, only convolution layers cannot model such spatial relations due to their local valid receptive field\footnote{Feature maps from deep CNNs like ResNet usually have large receptive fields due to deep architectures, whereas the study of~\cite{peng2017large} has shown that CNNs are apt to extract information mainly from smaller regions in receptive fields, which are called valid receptive fields.}.
\par
Nevertheless, under some circumstances, spatial relations are of paramount importance, particularly when a region in an image exhibits significant visual ambiguities. To address this issue, several attempts have been made to introduce spatial relations into networks by using either graphical models or spatial propagation networks. However, these methods seek to capture global spatial relations implicitly with a chain propagation way, whose effectiveness depends heavily on the learning effect of long-term memorization. Consequently, these models may not work well in some cases like aerial scenes (see Figure~\ref{fig:visual1} and Figure~\ref{fig:visual2}), in which long-range spatial relations often exist (\textit{cf}. Figure~\ref{fig:fig1}). Hence, explicit modeling of long-range relations may provide additional crucial information but still remains underexplored for semantic segmentation.
\par
This work is inspired by the recent success of relation networks in visual question answering~\cite{santoro2017simple}, object detection~\cite{relationdet}, and activity recognition in videos~\cite{zhou2017temporal}. Being able to reason about relationships between entities is momentous for intelligent decision-making. A relation network is capable of inferring relationships between an individual entity (\eg, a patch in an image) and a set of other entities (\eg, all patches in the image) by agglomerating information. The relations vary at both long-range and short-range scales and are learned automatically, driven by tasks. Moreover, a relation network can model dependencies between entities, without making excessive assumptions on their feature distributions and locations.
\par
In this work, our goal is to increase the representation capacity of a fully convolutional network (FCN) for semantic segmentation in aerial scenes by using relation modules: describing relationships between observations in convolved images and producing relation-augmented feature representations. Given that convolutions operate by blending spatial and cross-channel information together, we capture relations in both spatial and channel domains. More specifically, two plug-and-play modules---a spatial relation module and a channel relation module---are appended on top of feature maps of an FCN to learn different aspects of relations and then generate spatial relation-augmented and channel relation-augmented features, respectively, for semantic segmentation. By doing so, relationships between any two spatial positions or feature maps can be modeled and used to further enhance feature representations. Furthermore, we study empirically two ways of integrating two relation modules---serial and parallel. 
\par
\vspace{0.5em}
\noindent\textbf{Contributions.} This work's contributions are threefold.
\begin{itemize}
  \item We propose a simple yet effective and interpretable relation-augmented network that enables spatial and channel relational reasoning in networks for semantic segmentation on aerial imagery.
  \item A spatial relation module and a channel relation module are devised to explicitly model global relations, which are subsequently harnessed to produce spatial- and channel-augmented features.
  \item We validate the effectiveness of our relation modules through extensive ablation studies.
\end{itemize}
\begin{figure*}[ht]
    \centering
    \includegraphics[width=0.9\textwidth]{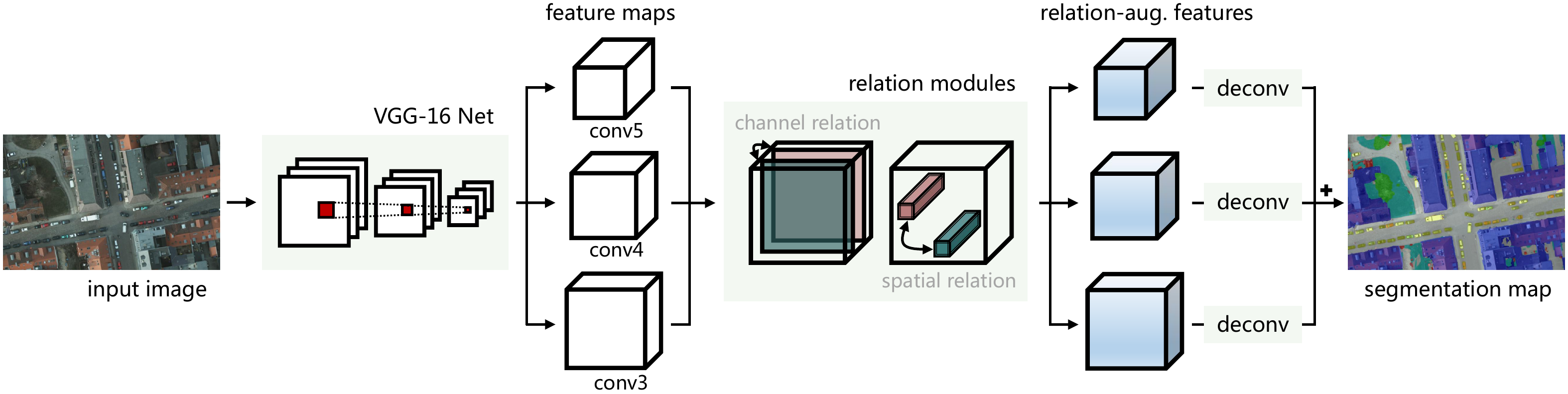}
    \caption{An overview of the relation module-equipped fully convolutional network.}
    \label{fig:net}
\end{figure*}
\section{Related Work}
\label{sec:related}
\noindent\textbf{Semantic segmentation of aerial imagery.} Earlier studies~\cite{Tokarczyk15} have focused on extracting useful low-level, handcrafted visual features and/or modeling mid-level semantic features on local portions of images (\cite{Kaiser17,Marcoscvpr18,torontocity,marcos2018land,Marmanis18,zhu2017deep,hua2019recurrently} employ deep CNNs and have made a great leap towards end-to-end aerial image parsing. In addition, there are numerous contests aiming at semantic segmentation from overhead imagery recently, \eg, Kaggle\footnote{\url{https://www.kaggle.com/c/dstl-satellite-imagery-feature-detection}}, SpaceNet\footnote{\url{https://spacenetchallenge.github.io/}}, and DeepGlobal\footnote{\url{http://deepglobe.org/challenge.html}}.
\par
\vspace{0.5em}
\noindent\textbf{Graphical models.} There are many graphical model-based methods being employed to achieve better semantic segmentation results. For example, the work in~\cite{deeplab} makes use of a CRF as post-processing to improve the performance of semantic segmentation. \cite{crfasrnn} and \cite{deepparsingnet} further make the CRF module differentiable and integrate it as a joint-trained part within networks. Moreover, low-level visual cues, \eg, object contours, have also been considered structure information~\cite{Bertasius16,Chen16}. These approaches, however, are sensitive to changes in appearance and expensive due to iterative inference processes required.
\par
\vspace{0.5em}
\noindent\textbf{Spatial propagation networks.} Learning spatial propagation with networks for semantic segmentation have attracted high interests in recent years. In~\cite{Maire16}, the authors try to predict entities of an affinity matrix directly by learning a CNN, which presents a good segmentation performance, while the affinity is followed by a nondifferentiable solver for spectral embedding, which results in the fact that the whole model cannot be trained end-to-end. The authors of~\cite{Liunips17} train a CNN model to learn a task-dependent affinity matrix by converting the modeling of affinity to learning a local linear spatial propagation. Several recent works~\cite{Keeccv18,Liueccv18,Chengeccv18} focus on the extension of this work. In~\cite{Bell16,pan2017spatial}, spatial relations are modeled and reinforced via interlayer propagation. \cite{Bell16} proposes an Inside-Outside Net (ION) where four independent recurrent networks that move in four directions are used to pass information along rows or columns. \cite{pan2017spatial} utilizes four slice-by-slice convolutions within feature maps, enabling message passings between neighboring rows and columns in a layer. The spatial propagation of these methods is serial in nature, and thus each position could only receive information from its neighbors.
\par
\vspace{0.5em}
\noindent\textbf{Relation networks.} Recently, the authors of~\cite{santoro2017simple} propose a relational reasoning network for the problem of visual question answering, and this network achieves a super-human performance. Later, \cite{zhou2017temporal} proposes a temporal relation network to enable multi-scale temporal relational reasoning in networks for video classification tasks. In~\cite{relationdet}, the authors propose an object relation module, which allows modeling relationships among sets of objects, for object detection tasks. Our work is motivated by the recent success of these works, but we focus on modeling spatial and channel relations in a CNN for semantic segmentation.
\par
\vspace{1em}
Unlike graphical model-based \cite{koller, Wainwright} and spatial propagation network-based methods, we explicitly take spatial relations and channel relations into account, so that semantic image segmentation could beneﬁt from short- and long-range relational reasoning.
\par
\section{Our Approach}
\label{sec:method}
In this section, an overview of the proposed relational context-aware network is given to present a comprehensive picture. Afterwards, two key components, the spatial relation module and the channel relation module, are introduced, respectively. Finally, we describe the strategy of integrating these modules for semantic segmentation.

\subsection{Overview}
As illustrated in Fig.~\ref{fig:net}, the proposed network takes VGG-16 \cite{vgg} as a backbone to extract multi-level features. Outputs of \textit{conv3}, \textit{conv4}, and \textit{conv5} are fed into the channel and spatial relation modules (see Figure~\ref{fig:net}) for generating relation-augmented features. These features are subsequently fed into respective convolutional layers with $1\times 1$ filters to squash the number of channels to the number of categories. Finally, the convolved feature maps are upsampled to a desired full resolution and element-wise added to generate final segmentation maps.

\subsection{Spatial Relation Module}
In order to capture global spatial relations, we employ a spatial relation module, where the spatial relation is defined as a composite function with the following equation:
\begin{equation}
\label{eq:sr}
    {\rm SR}(\bm{x}_i,\bm{x}_j) = f_{\phi_s}(g_{\theta_s}(\bm{x}_i,\bm{x}_j))\,.
\end{equation}

Denote by $\bm{X}\in\mathbb{R}^{C\times H\times W}$ a random variable representing a set of feature maps. $\bm{x}_i$ and $\bm{x}_j$ are two feature-map vectors and identified by spatial positions indices $i$ and $j$. The size of $\bm{x}_i$ and $\bm{x}_j$ is $C\times 1\times 1$. To model a compact relationship between these two feature-map vectors, we make use of an embedding dot production as $g_{\theta_s}$ instead of a multilayer perceptron (MLP), and the latter is commonly used in relational reasoning modules \cite{santoro2017simple, zhou2017temporal}. Particularly, $g_{\theta_s}$ is defined as follows:
\begin{equation}
\label{eq:dot1}
    g_{\theta_s}(\bm{x}_i,\bm{x}_j)=u_s(\bm{x}_i)^Tv_s(\bm{x}_j)\,,
\end{equation}
where $u_s(\bm{x}_i)=\bm{W}_{u_s}\bm{x}_i$ and $v_s(\bm{x}_j)=\bm{W}_{v_s}\bm{x}_j$. $\bm{W}_{u_s}$ and $\bm{W}_{v_s}$ are weight matrices and can be learned during the training phase. Considering computational efficiency, we realize Eq.~(\ref{eq:dot1}) in matrix format with the following steps:
\begin{enumerate}
  \setlength{\itemsep}{0pt}
  \setlength{\parsep}{0pt}
  \setlength{\parskip}{0pt}
  \item Feature maps $\bm{X}$ are fed into two convolutional layers with $1\times 1$ filters to generate $u_s(\bm{X})$ and $v_s(\bm{X})$, respectively.
  \item Then $u_s(\bm{X})$ and $v_s(\bm{X})$ are reshaped (and transposed) into $HW\times C$ and $C\times HW$, correspondingly.
  \item Eventually, the matrix multiplication of $u_s(\bm{X})$ and $v_s(\bm{X})$ is conducted to produce a $HW\times HW$ matrix, which is further reshaped to form a spatial relation feature of size $HW\times H\times W$.
\end{enumerate}

It is worth nothing that the spatial relation feature is not further synthesized (\eg, summed up), as fine-grained contextual characteristics are essential in semantic segmentation tasks. Afterwards, we select the ReLU function as $f_{\phi_s}$ to eliminate negative spatial relations.

However, relying barely on spatial relations leads to a partial judgment. Therefore, we further blend the spatial relation feature and original feature maps $\bm{X}$ as follows:
\begin{equation}
\label{eq:sr3}
    \bm{X}_s=[\bm{X}, {\rm SR}(\bm{X})]\,.
\end{equation}

Here we simply use a concatenation operation, i.e., $[\cdot,\cdot]$, to enhance original features with spatial relations. By doing so, output features are abundant in global spatial relations, while high-level semantic features are also preserved.
\par
\subsection{Channel Relation Module}
Although the spatial relation module is capable of capturing global contextual dependencies for identifying various objects, misdiagnoses happen when objects share similar distribution patterns but vary in channel dimensionality. In addition, a recent work~\cite{SENet} has shown the benefit of enhancing channel encoding in a CNN for image classification tasks. Therefore, we propose a channel relation module to model channel relations, which can be used to enhance feature discriminabilities in the channel domain. Similar to the spatial relation module, we define the channel relation as a composite function with the following equation:
\begin{equation}
\label{eq:cr}
    {\rm CR}(\bm{X}_p,\bm{X}_q) = f_{\phi_c}(g_{\theta_c}(\bm{X}_p,\bm{X}_q))\,,
\end{equation}
where the input is a set of feature maps $\bm{X}=\{\bm{X}_1,\bm{X}_2,\cdots,\bm{X}_C\}$, and $\bm{X}_p$ as well as $\bm{X}_q$ represents the $p$-th and the $q$-th channels of $\bm{X}$. Embedding dot production is employed to be $g_{\theta_c}$, defined as
\begin{equation}
\label{eq:dot2}
    g_{\theta_c}(\bm{X}_p, \bm{X}_q)=u_c({\rm GAP}(\bm{X}_p))^Tv_c({\rm GAP}(\bm{X}_q))\,,
\end{equation}
for capturing global relationships between feature map pairs, where ${\rm GAP}(\cdot)$ denotes the global average pooling function. Notably, considering that the preservation of spatial structural information distracts the analysis of channel inter-dependencies, we adopt averages of $\bm{X}_p$ and $\bm{X}_q$ as channel descriptors before performing dot production. More specifically, we feed feature maps into a global average pooling layer for generating a set of channel descriptors of size $C\times 1\times 1$, and then exploit two convolutional layers with $1\times 1$ filters to produce $u_c(\bm{X})$ and $v_c(\bm{X})$, respectively. Afterwards, an outer production is performed to generate a $C\times C$ channel relation feature, where the element located at $(p,q)$ indicates $g_{\theta_c}(\bm{X}_p,\bm{X}_q)$.

Furthermore, we emphasize class-relevant channel relations as well as suppress irrelevant channel dependencies by adopting a softmax function as $f_{\phi_c}$, formulated as
\begin{equation}
\label{eq:softmax}
    f_{\phi_c}(g_{\theta_c}(\bm{X}_p, \bm{X}_q))=\frac{\exp(g_{\theta_c}(\bm{X}_p, \bm{X}_q))}{\sum_{q=1}^{C}\exp(g_{\theta_c}(\bm{X}_p, \bm{X}_q))}\,,
\end{equation}
where we take $\bm{X}_p$ as an example. Consequently, a discriminative channel relation map ${\rm CR}(\bm{X})$ can be obtained, where each element represents the corresponding pairwise channel relation.

To integrate ${\rm CR}(\bm{X})$ and original feature maps $\bm{X}$, we reshape $\bm{X}$ into a matrix of $C \times HW$ and employ a matrix multiplication as follows:
\begin{equation}
\label{eq:sr3}
    \bm{X}_c=\bm{X}^{T}{\rm CR}(\bm{X})\,.
\end{equation}

With this design, the input features are enhanced with channel relations and embedded with not only initial discriminative channel properties but also global inter-channel correlations. Eventually, $\bm{X}_c$ is reshaped to $C\times H\times W$ and fed into subsequent procedures.
\par

\subsection{Integration of Relation Modules}
\begin{figure}[t]
    \centering
    \subfloat[]{\includegraphics[width=0.9\textwidth]{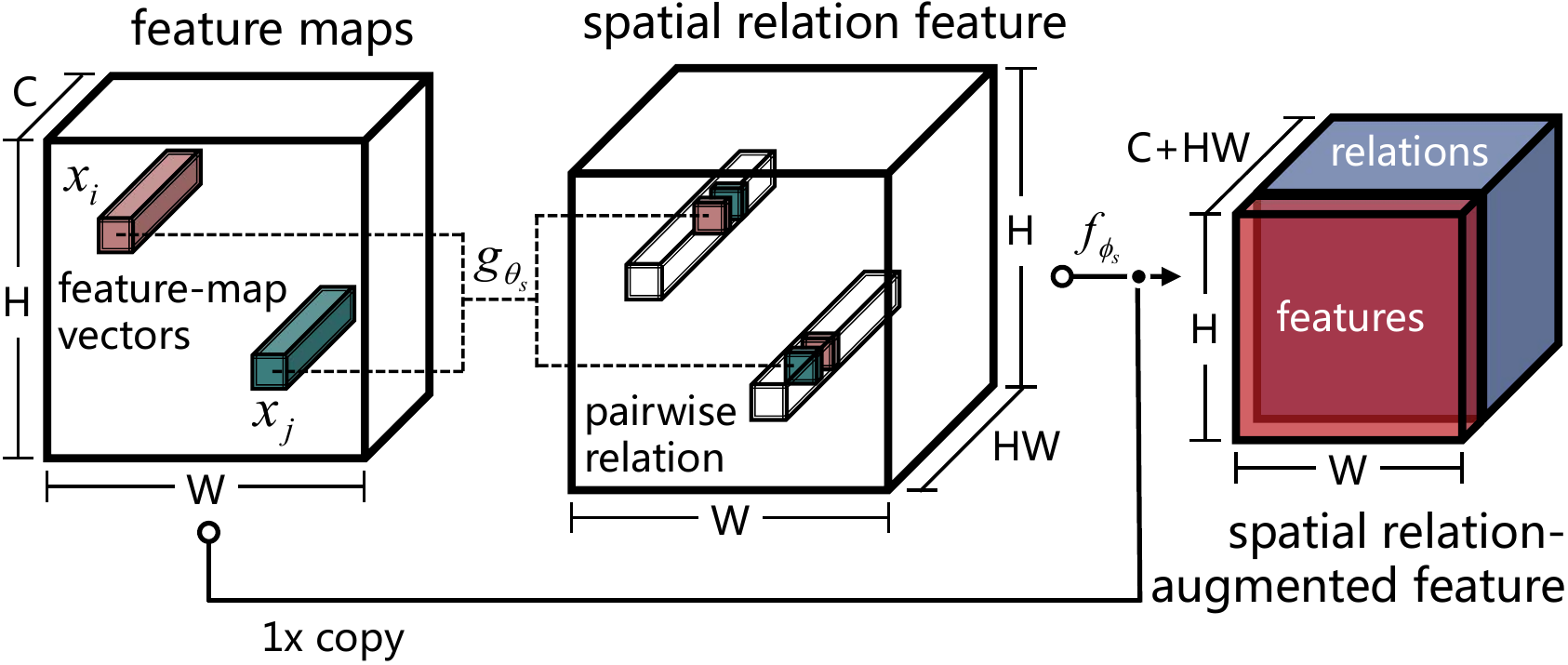}}
    \vfill
    \subfloat[]{\includegraphics[width=0.9\textwidth]{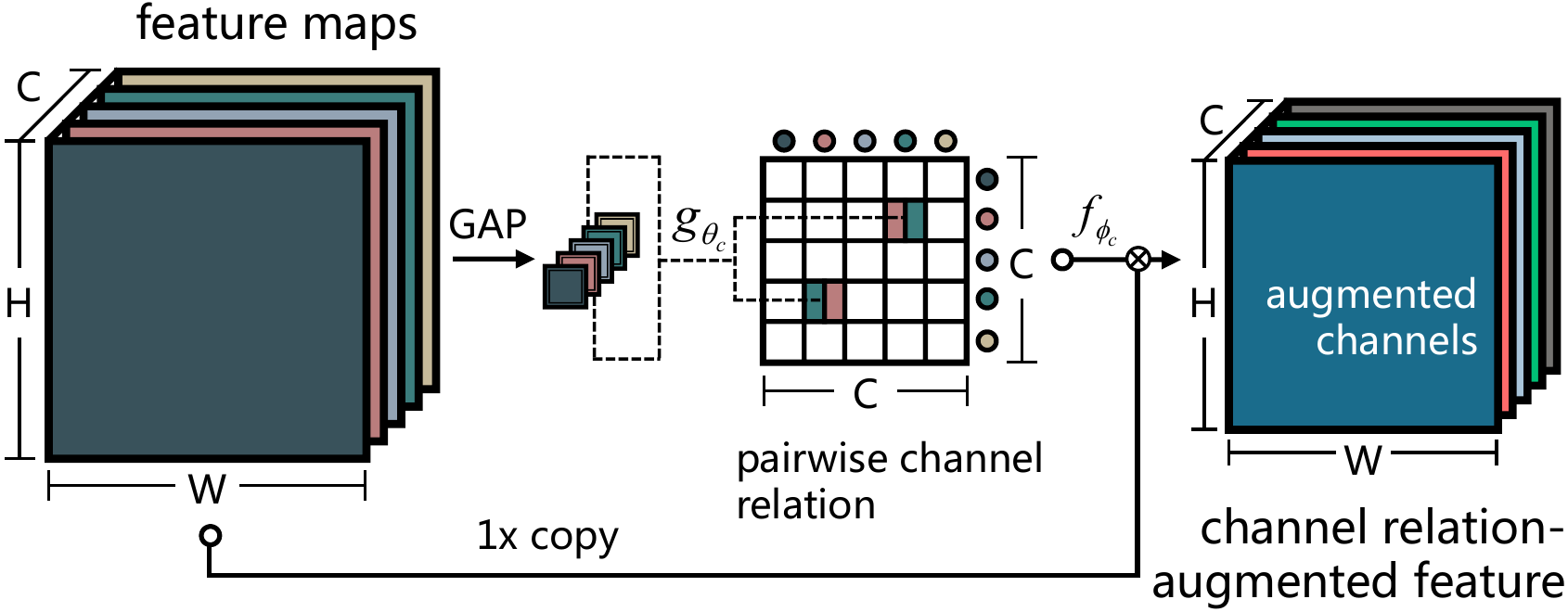}}
    \renewcommand{\figurename}{Fig}
    \caption{Diagrams of (a) spatial relation module and (b) channel relation module.}
    \label{fig:srm}
\end{figure}

\begin{figure}[t]
\centering
\subfloat[]{
\includegraphics[width=0.75\textwidth]{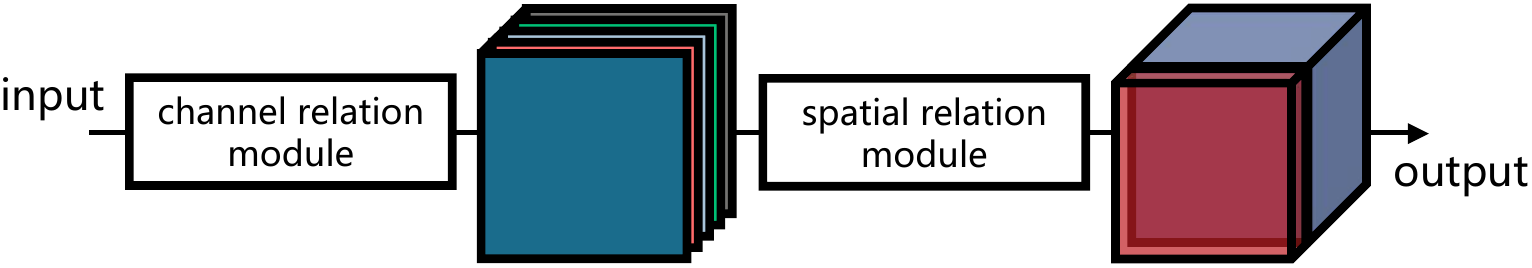}}\\
\subfloat[]{
\includegraphics[width=0.7\textwidth]{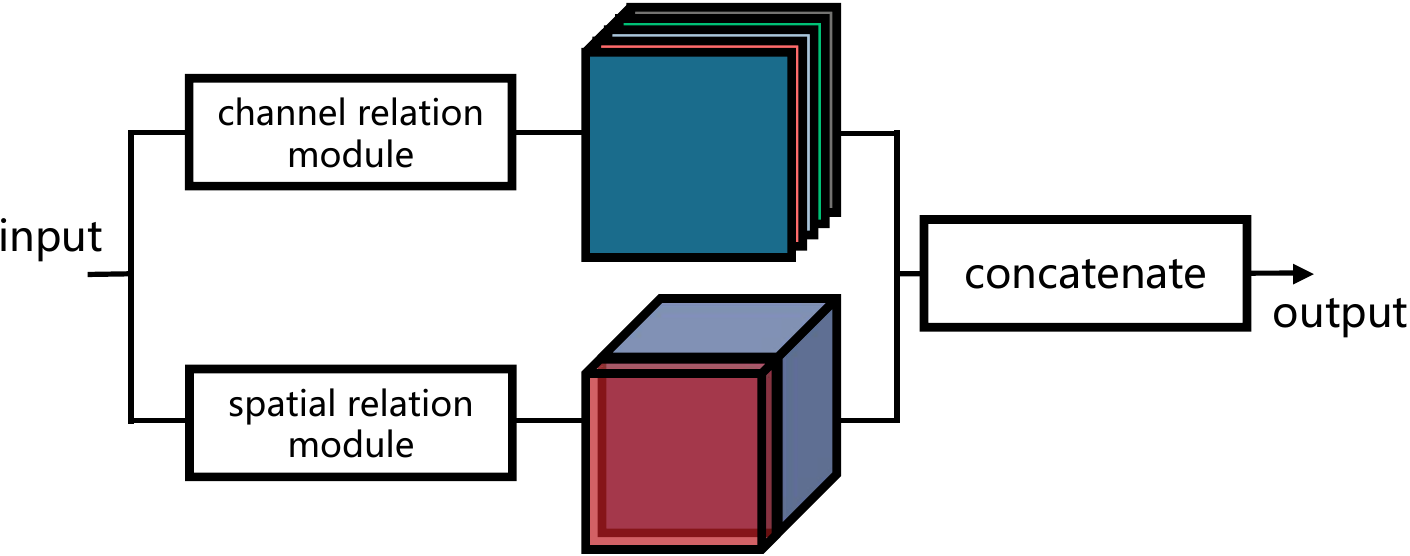}}
\renewcommand{\figurename}{Fig}
\caption{\label{fig:serial_parallel} Two integration manners: (a) serial and (b) parallel.}
\end{figure}

In order to jointly enjoy benefits from spatial and channel relation modules, we further aggregate features $\bm{X}_s$ and $\bm{X}_c$ to generate spatial and channel relation-augmented features. As shown in Fig.~\ref{fig:serial_parallel}, we investigate two integration patterns, namely serial integration and parallel integration, to blend $\bm{X}_s$ and $\bm{X}_c$. For the former, we append the spatial relation module to the channel relation module and infer $\bm{X}_s$ from $\bm{X}_c$ instead of $\bm{X}$, as presented in Eq.~(\ref{eq:sr}) and Eq.~(\ref{eq:sr3}). For the latter, spatial relation-augmented features and channel relation-augmented features are obtained simultaneously and then aggregated by performing concatenation. Influences of different strategies are discussed in Section \ref{chap:ablation}.

\section{Experiments}
\label{sec:exp}
To verify the effectiveness of long-range relation modeling in our network, aerial image datasets are used in experiments. This is because aerial images are taken from nadir view, and the spatial distribution/relation of objects in these images is diverse and complicated, as shown in Figure~\ref{fig:fig1}. Thus, we perform experiments on two aerial image semantic segmentation datasets, \ie, ISPRS Vaihingen and Potsdam datasets, and results are discussed in subsequent sections.

\subsection{Experimental Setup}
\noindent\textbf{Datasets.} The Vaihingen dataset\footnote{\url{http://www2.isprs.org/commissions/comm3/wg4/2d-sem-label-vaihingen.html}} is composed of 33 aerial images collected over a 1.38 km\textsuperscript{2} area of the city, Vaihingen, with a spatial resolution of 9 cm. The average size of each image is $2494 \times 2064$ pixels, and each of them has three bands, corresponding to near infrared (NIR), red (R), and green (G) wavelengths. Notably, DSMs, which indicate the height of all object surfaces in an image, are also provided as complementary data. Among these images, 16 of them are manually annotated with pixel-wise labels, and each pixel is classified into one of six land cover classes. Following the setup in \cite{Maggiori2017HighResolutionAI,volpi2017dense,sherrah2016fully,marcos2018land}, we select 11 images (image IDs: 1, 3, 5, 7, 13, 17, 21, 23, 26, 32, 37) for training, and the remaining five images (image IDs: 11, 15, 28, 30, 34) are used to test our model.
\par
The Potsdam dataset\footnote{\url{http://www2.isprs.org/commissions/comm3/wg4/2d-sem-label-potsdam.html}} consists of 38 high resolution aerial images, which covers an area of 3.42 km\textsuperscript{2}, and each aerial image is captured in four channels (NIR, R, G, and blue (B)). The size of all images is $6000 \times 6000$ pixels, which are annotated with pixels-level labels of six classes as the Vaihingen dataset. The spatial resolution is 5 cm, and coregistered DSMs are available as well. To train and evaluate networks, we utilize 17 images (image IDs: 02\_10, 03\_10, 03\_11, 03\_12, 04\_11, 04\_12, 05\_10, 05\_12, 06\_08, 06\_09, 06\_10, 06\_11, 06\_12, 07\_07, 07\_09, 07\_11, 07\_12) for training and build the test set with the remaining images (image IDs: 02\_11, 02\_12, 04\_10, 05\_11, 06\_07, 07\_08, 07\_10), which follows the setup in \cite{Maggiori2017HighResolutionAI, sherrah2016fully}.
\par
\vspace{0.5em}
\noindent\textbf{Implementation.} The proposed network is initialized with separate strategies with respect to two dominant components: the feature extraction module is initialized with CNNs pre-trained on ImageNet dataset~\cite{imagenet_cvpr09}, while convolutional layers in relation modules are initialized with a Glorot uniform initializer. Notably, weights in the feature extraction module are trainable and fine-tuned during the training phase.
\par
Regarding the used optimizer, we choose Nestrov Adam~\cite{nadam2} and set parameters of the optimizer as recommended: $\beta_1=0.9$, $\beta_2=0.999$, and $\epsilon=1\mathrm{e}{-08}$. The learning rate is initialized as $2\mathrm{e}{-04}$ and decayed by 0.1 when validation loss is saturated. The loss of our network is simply defined as categorical cross-entropy. We implement the network on TensorFlow and train it on one NVIDIA Tesla P100 16GB GPU for 250k iterations. The size of the training batch is 5, and we stop training when the validation loss fails to decrease.
\par
\vspace{0.5em}
\noindent\textbf{Evaluation metric.} To evaluate the performance of networks, we calculate $F_1$ score with the following formula:
\begin{equation}
\label{eq:f1}
F_1 = (1+\beta^2)\cdot\frac{precision \cdot recall}{\beta^2\cdot precision+recall}, \hspace{1em} \beta=1,
\end{equation}
for each category. Furthermore, mean $F_1$ score is computed by averaging all $F_1$ scores to assess models impartially. Notably, a large $F_1$ score suggests a better result. Besides, mean IoU (mIoU) and overall accuracy (OA) that indicates overall pixel accuracy, are also calculated for a comprehensive comparison with different models.

\subsection{An Ablation Study for Relation Modules}
\label{chap:ablation}

\begin{table}[!t]
\centering
\begin{threeparttable}
\caption{Ablation Study on the Vaihingen Dataset.}
\label{tab:setting}
\begin{tabular}{c|cc|c|c}
\Xhline{2\arrayrulewidth}
Model Name & crm & srm & mean $F_1$ & OA \\
\hline
\hline
Baseline FCN~\cite{long2015fully} &  &  & 83.74 & 86.51 \\
RA-FCN-crm & \checkmark &  & 87.24 & 88.38  \\
RA-FCN-srm & & \checkmark & 88.36 & 89.03 \\
\hline
P-RA-FCN & \checkmark & \checkmark & 88.50 & 89.18 \\
S-RA-FCN & \checkmark & \checkmark & \textbf{88.54} & \textbf{89.23} \\
\Xhline{2\arrayrulewidth}
\end{tabular}
\begin{tablenotes}
\item[1] RA-FCN indicates the proposed relation-augmented FCN.
\item[2] crm indicates the channel relation module.
\item[3] srm indicates the spatial relation module.
\item[4] P-RA-FCN indicates that crm and srm are appended on top of the backbone in parallel.
\item[5] S-RA-FCN indicates that crm is followed by srm.
\end{tablenotes}
\end{threeparttable}
\end{table}
\begin{table*}[!t]
\caption{Experimental Results on the Vaihingen Dataset}
\label{tab:vaihingen}
\centering
\begin{tabular}{l|ccccc|c|c|c}
\Xhline{2\arrayrulewidth}
Model Name & Imp. surf. & Build. & Low veg. & Tree & Car & mean $F_1$ & mIoU & OA \\
\hline
\hline
SVL-boosting+CRF$^{*}$~\cite{svl_5} & 86.10 & 90.90 & 77.60 & 84.90 & 59.90 & 79.90 & - & 84.70\\ 
RF+dCRF$^{*}$~\cite{hust} & 86.90 & 92.00 & 78.3 & 86.90 & 29.00 & 74.60 & - & 85.90\\
CNN-FPL$^{*}$~\cite{volpi2017dense} & - & - & - & - & - & 83.58 & - & 87.83\\
FCN~\cite{long2015fully} & 88.67 & 92.83 & 76.32 & 86.67 & 74.21 & 83.74 & 72.69 & 86.51 \\
FCN-dCRF~\cite{deeplab} & 88.80 & 92.99 & 76.58 & 86.78 & 71.75 & 83.38 & 72.28 & 86.65 \\
SCNN~\cite{pan2017spatial} & 88.21 & 91.80 & 77.17 & 87.23 & 78.60 & 84.40 & 73.73 & 86.43 \\
Dilated FCN~\cite{deeplab} & 90.19 & 94.49 & 77.69& 87.24 & 76.77& 85.28 & - & 87.70 \\
FCN-FR$^{*}$~\cite{Maggiori2017HighResolutionAI} & \textbf{91.69} & \textbf{95.24} & 79.44 & 88.12 & 78.42 & 86.58 & - & 88.92\\
PSPNet (VGG16)~\cite{cas_l4} & 89.92 & 94.36 & 78.19 & 87.12 & 72.97 & 84.51 & 73.97 & 87.62\\
RotEqNet$^{*}$~\cite{marcos2018land} & 89.50 & 94.80 & 77.50 & 86.50 & 72.60 & 84.18 & - & 87.50\\
\hline
RA-FCN-srm & 91.01 & 94.86 & 80.01 & 88.74 & 87.16 & 88.36 & 79.48 & 89.03 \\
P-RA-FCN & 91.46 & 95.02 & 80.40 & 88.56 & \textbf{87.08} & 88.50 & 79.72 & 89.18 \\
\textbf{S-RA-FCN} & 91.47 & 94.97 & \textbf{80.63} & \textbf{88.57} & 87.05 & \textbf{88.54} & \textbf{79.76} & \textbf{89.23} \\
\Xhline{2\arrayrulewidth}
\end{tabular}
\end{table*}
\begin{figure*}[!t]
\captionsetup[subfigure]{labelformat=empty, captionskip=2pt}
\centering
\subfloat{\includegraphics[width=0.112\textwidth]{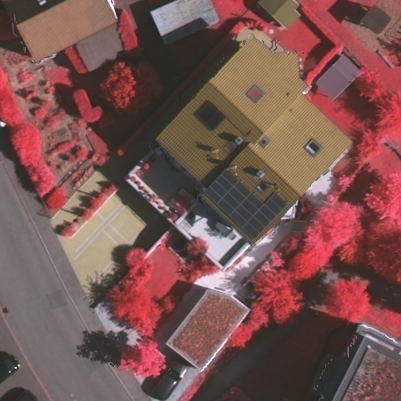}}
\hspace{-0.1em}
\subfloat{\includegraphics[width=0.112\textwidth]{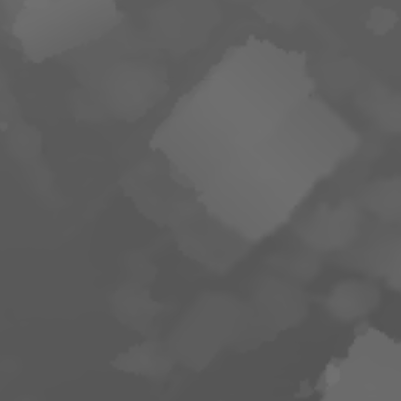}}
\hspace{-0.1em}
\subfloat{\includegraphics[width=0.112\textwidth]{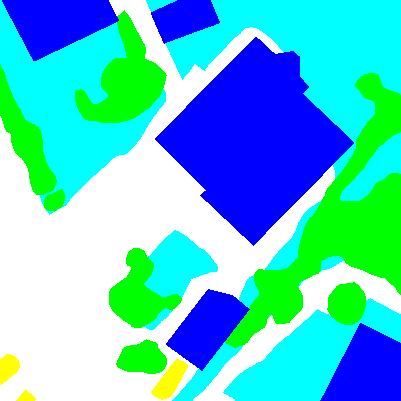}}
\hspace{-0.1em}
\subfloat{\includegraphics[width=0.112\textwidth]{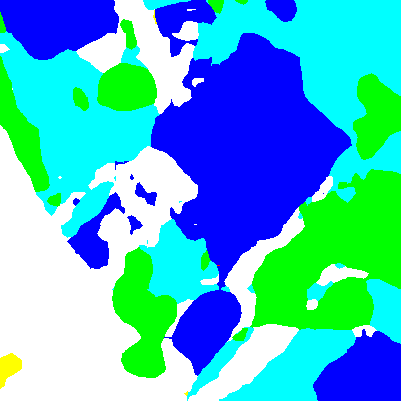}}
\hspace{-0.1em}
\subfloat{\includegraphics[width=0.112\textwidth]{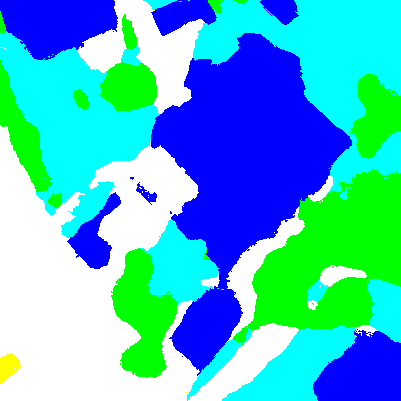}}
\hspace{-0.1em}
\subfloat{\includegraphics[width=0.112\textwidth]{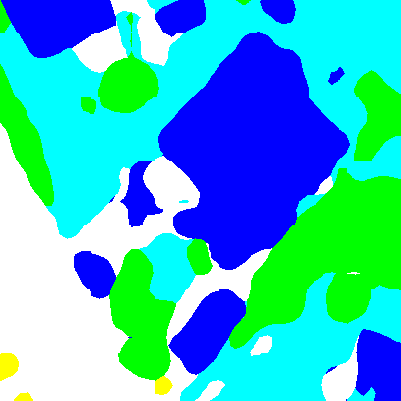}}
\hspace{-0.1em}
\subfloat{\includegraphics[width=0.112\textwidth]{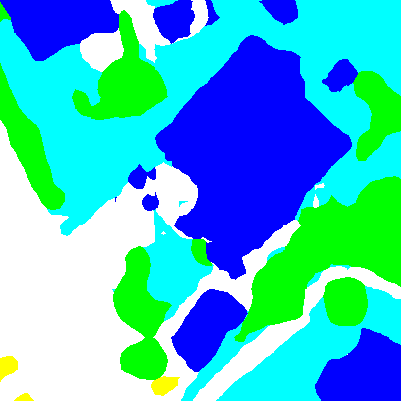}}
\hspace{-0.1em}
\subfloat{\includegraphics[width=0.112\textwidth]{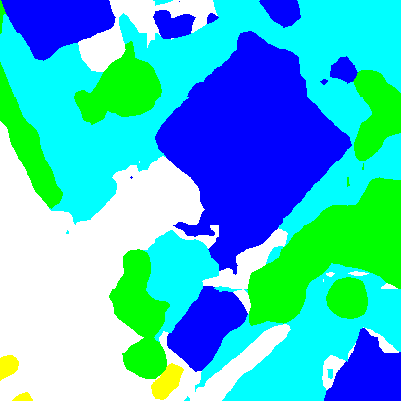}}
\vspace{-0.8em}

\subfloat{\includegraphics[width=0.112\textwidth]{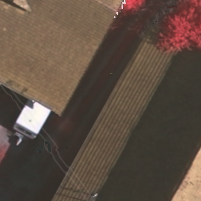}}
\hspace{-0.1em}
\subfloat{\includegraphics[width=0.112\textwidth]{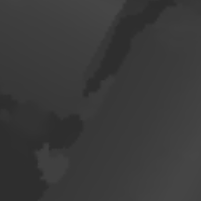}}
\hspace{-0.1em}
\subfloat{\includegraphics[width=0.112\textwidth]{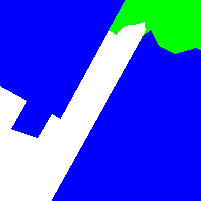}}
\hspace{-0.1em}
\subfloat{\includegraphics[width=0.112\textwidth]{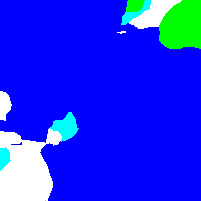}}
\hspace{-0.1em}
\subfloat{\includegraphics[width=0.112\textwidth]{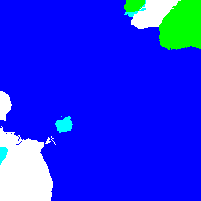}}
\hspace{-0.1em}
\subfloat{\includegraphics[width=0.112\textwidth]{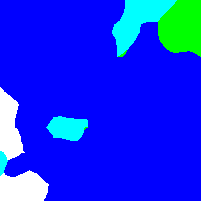}}
\hspace{-0.1em}
\subfloat{\includegraphics[width=0.112\textwidth]{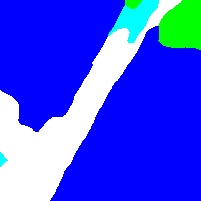}}
\hspace{-0.1em}
\subfloat{\includegraphics[width=0.112\textwidth]{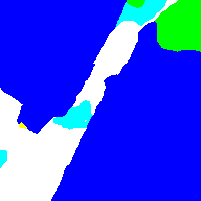}}
\vspace{-0.8em}

\subfloat{\includegraphics[width=0.112\textwidth]{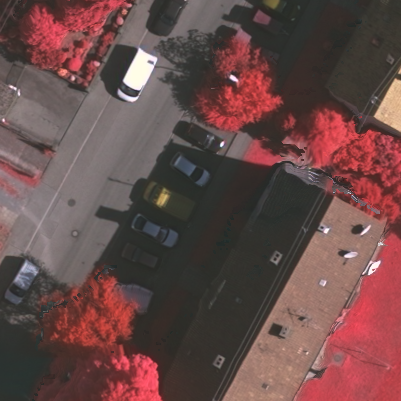}}
\hspace{-0.1em}
\subfloat{\includegraphics[width=0.112\textwidth]{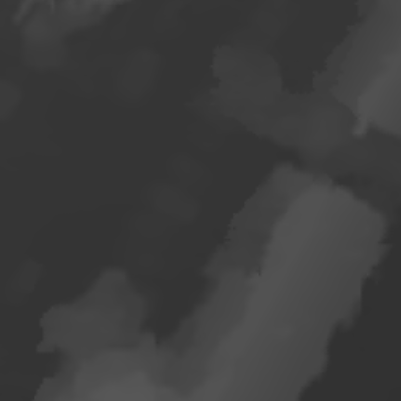}}
\hspace{-0.1em}
\subfloat{\includegraphics[width=0.112\textwidth]{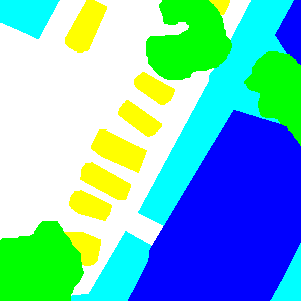}}
\hspace{-0.1em}
\subfloat{\includegraphics[width=0.112\textwidth]{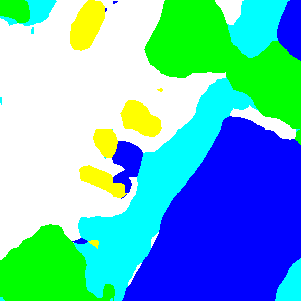}}
\hspace{-0.1em}
\subfloat{\includegraphics[width=0.112\textwidth]{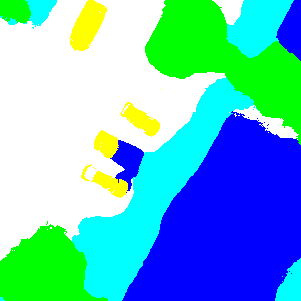}}
\hspace{-0.1em}
\subfloat{\includegraphics[width=0.112\textwidth]{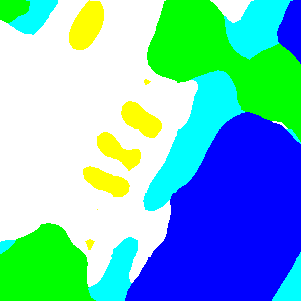}}
\hspace{-0.1em}
\subfloat{\includegraphics[width=0.112\textwidth]{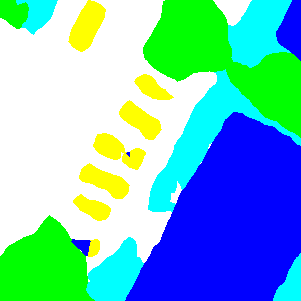}}
\hspace{-0.1em}
\subfloat{\includegraphics[width=0.112\textwidth]{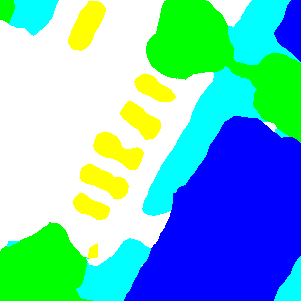}}
\vspace{-0.8em}

\subfloat[Image]{\includegraphics[width=0.112\textwidth]{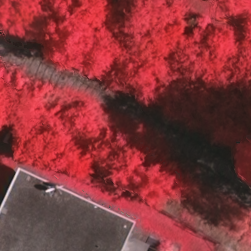}}
\hspace{-0.1em}
\subfloat[nDSM]{\includegraphics[width=0.112\textwidth]{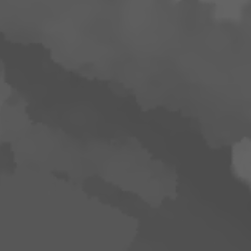}}
\hspace{-0.1em}
\subfloat[Ground Truth]{\includegraphics[width=0.112\textwidth]{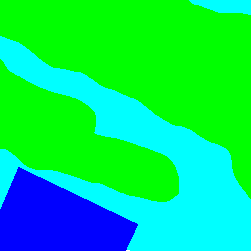}}
\hspace{-0.1em}
\subfloat[FCN]{\includegraphics[width=0.112\textwidth]{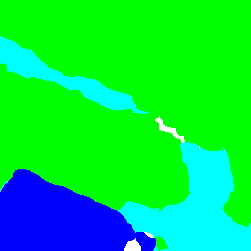}}
\hspace{-0.1em}
\subfloat[FCN-dCRF]{\includegraphics[width=0.112\textwidth]{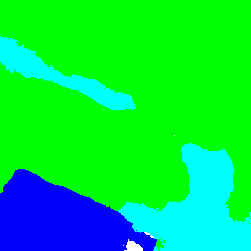}}
\hspace{-0.1em}
\subfloat[SCNN]{\includegraphics[width=0.112\textwidth]{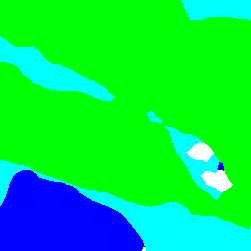}}
\hspace{-0.1em}
\subfloat[RA-FCN-srm]{\includegraphics[width=0.112\textwidth]{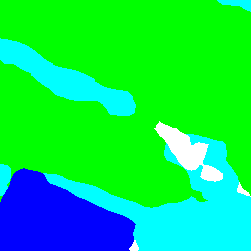}}
\hspace{-0.1em}
\subfloat[RA-FCN]{\includegraphics[width=0.112\textwidth]{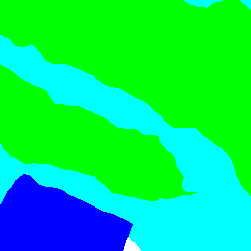}}
\caption{Examples of segmentation results on the Vaihingen dataset. Legend---white: impervious surfaces, \textcolor{blue}{blue}: buildings, \textcolor{cyan}{cyan}: low vegetation, \textcolor{green}{green}: trees, \textcolor{yellow}{yellow}: cars.}
\label{fig:visual1}
\end{figure*}

In our network, spatial and channel relation modules are employed to explore global relations in both spatial and channel domains. To validate the effectiveness of these modules, we perform ablation experiments (\textit{cf}. Table~\ref{tab:setting}). Particularly, instead of being utilized simultaneously, spatial and channel relation modules are embedded on top of the backbone (i.e., VGG-16), respectively. Besides, we also discuss different integration strategies (\ie, parallel and serial) of relation modules in Table~\ref{tab:setting}.
\par
The ablation experiments are conducted on the Vaihingen dataset. As can be seen in Table~\ref{tab:setting}, relation modules bring a significant improvement as compared to the baseline FCN (VGG-16), and various integration schemes lead to a slight influence on the performance of our network. In detailed, the use of only the channel relation module yields a result of 87.24\% in the mean $F_1$ score, which brings a 3.50\% improvement. Meanwhile, RA-FCN with only the spatial relation module outperforms the baseline by a 4.62\% gain in the mean $F_1$ score. In addition, we note that squeeze-and-excitation module~\cite{SENet} can also model dependencies between channels. However, in our experiments, the proposed channel relation module performs better.   
\par

\begin{table*}[t!]
\caption{Numerical Results on the Potsdam Dataset}
\label{tab:potsdam}
\centering
\begin{tabular}{l|cccccc|c|c|c}
\Xhline{2\arrayrulewidth}
Model Name & Imp. surf. & Build. & Low veg. & Tree & Car & Clutter & mean $F_1$ & mIoU & OA \\
\hline
\hline
FCN~\cite{long2015fully} & 88.61 & 93.29 & 83.29 & 79.83 & 93.02 & 69.77 & 84.63 & 78.34 & 85.59\\
FCN-dCRF~\cite{deeplab} & 88.62 & 93.29 & 83.29 & 79.83 & 93.03 & 69.79 & 84.64 & 78.35 & 85.60\\
SCNN~\cite{pan2017spatial} & 88.37 & 92.32 & 83.68 & 80.94 & 91.17 & 68.86 & 84.22 & 77.72 & 85.57 \\
Dilated FCN$^{*}$~\cite{deeplab} & 86.52 & 90.78 & 83.01 & 78.41 & 90.42 & 68.67 & 82.94 & - & 84.14\\
FCN-FR$^{*}$~\cite{Maggiori2017HighResolutionAI}
& 89.31 & 94.37 & 84.83 & 81.10 & 93.56 & 76.54 & 86.62 & - & 87.02\\
\hline
RA-FCN-srm & 90.48 & 93.74 & 85.67 & 83.10 & 94.34 & 74.02 & 86.89 & 81.23 & 87.61\\
P-RA-FCN & 90.92 & 94.20 & 86.64 & 83.00 & 94.44 & \textbf{77.88} & 87.85 & 81.85 & 88.30\\
\textbf{S-RA-FCN} & \textbf{91.33} & \textbf{94.70} & \textbf{86.81} & \textbf{83.47} & \textbf{94.52} & 77.27 & \textbf{88.01} & \textbf{82.38} & \textbf{88.59} \\
\Xhline{2\arrayrulewidth}
\end{tabular}
\end{table*}

\begin{figure*}[!t]
\captionsetup[subfigure]{labelformat=empty, captionskip=2pt}
\centering
\subfloat{\includegraphics[width=0.112\textwidth]{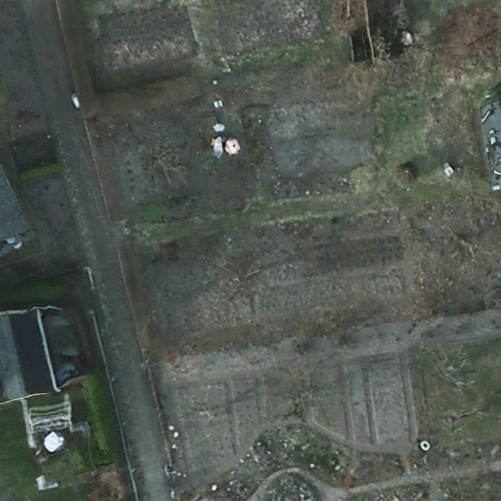}}
\hspace{-0.1em}
\subfloat{\includegraphics[width=0.112\textwidth]{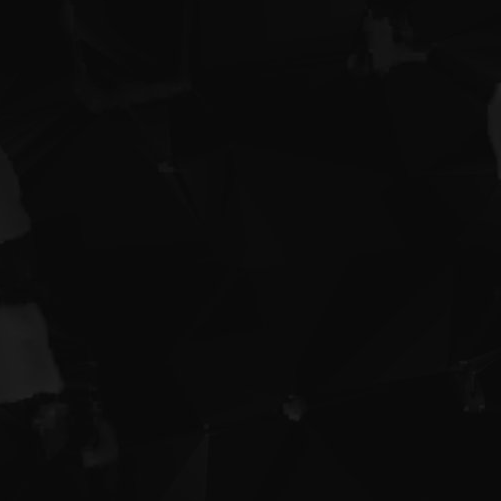}}
\hspace{-0.1em}
\subfloat{\includegraphics[width=0.112\textwidth]{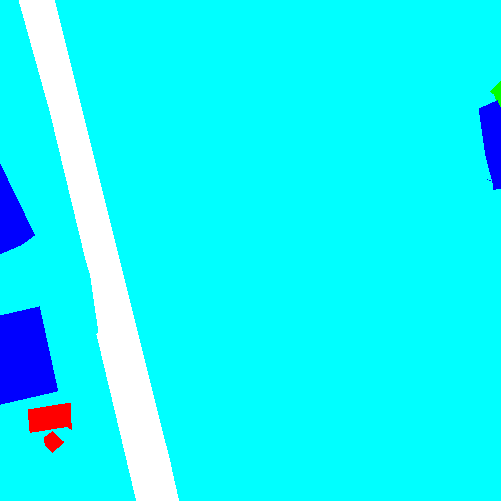}}
\hspace{-0.1em}
\subfloat{\includegraphics[width=0.112\textwidth]{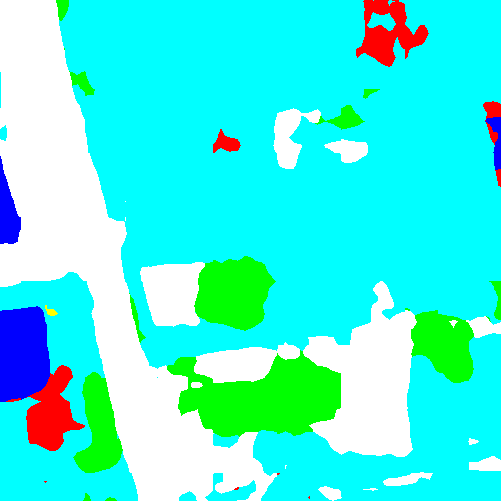}}
\hspace{-0.1em}
\subfloat{\includegraphics[width=0.112\textwidth]{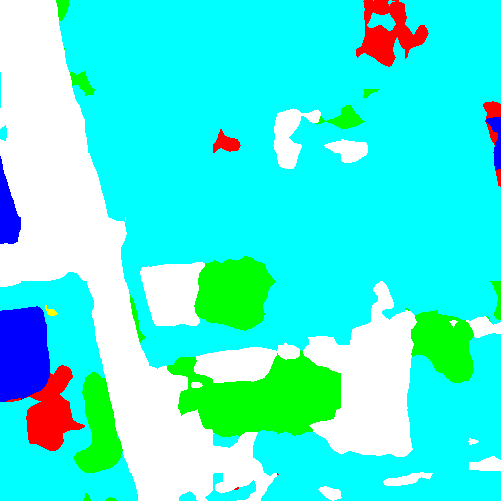}}
\hspace{-0.1em}
\subfloat{\includegraphics[width=0.112\textwidth]{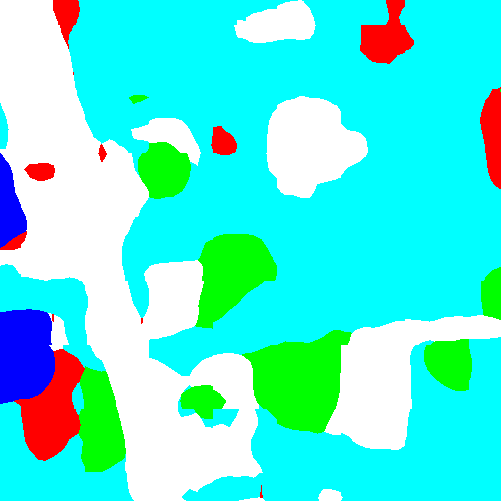}}
\hspace{-0.1em}
\subfloat{\includegraphics[width=0.112\textwidth]{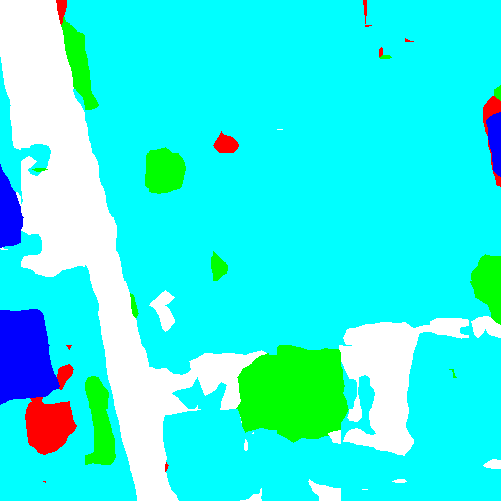}}
\hspace{-0.1em}
\subfloat{\includegraphics[width=0.112\textwidth]{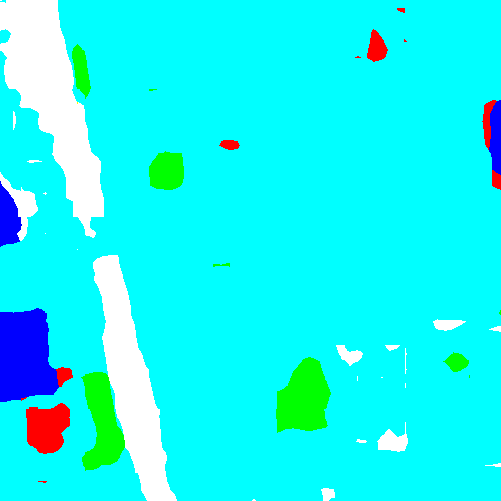}}
\vspace{-0.8em}

\subfloat{\includegraphics[width=0.112\textwidth]{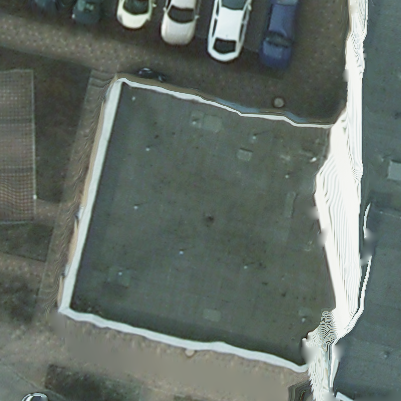}}
\hspace{-0.1em}
\subfloat{\includegraphics[width=0.112\textwidth]{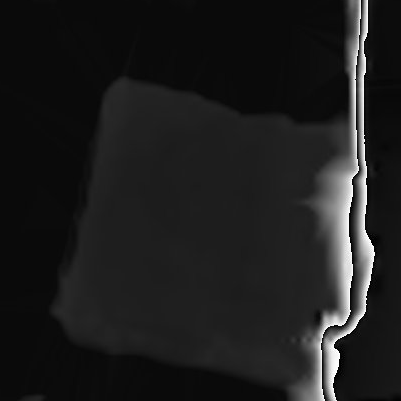}}
\hspace{-0.1em}
\subfloat{\includegraphics[width=0.112\textwidth]{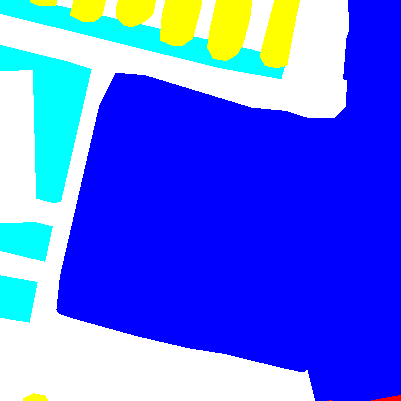}}
\hspace{-0.1em}
\subfloat{\includegraphics[width=0.112\textwidth]{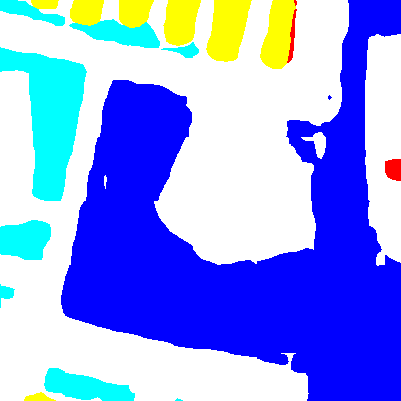}}
\hspace{-0.1em}
\subfloat{\includegraphics[width=0.112\textwidth]{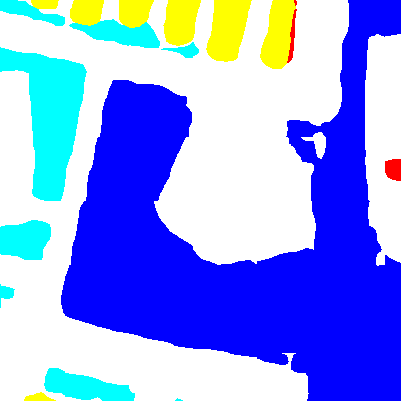}}
\hspace{-0.1em}
\subfloat{\includegraphics[width=0.112\textwidth]{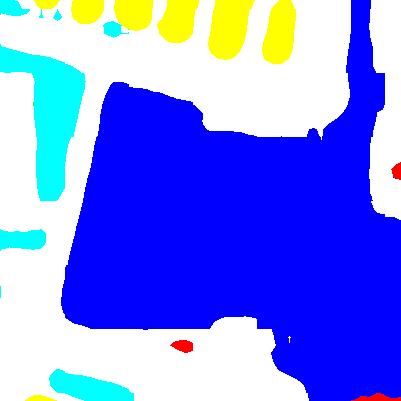}}
\hspace{-0.1em}
\subfloat{\includegraphics[width=0.112\textwidth]{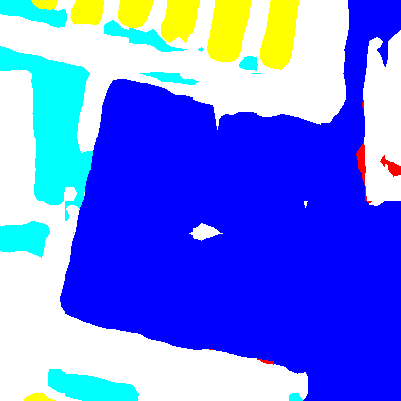}}
\hspace{-0.1em}
\subfloat{\includegraphics[width=0.112\textwidth]{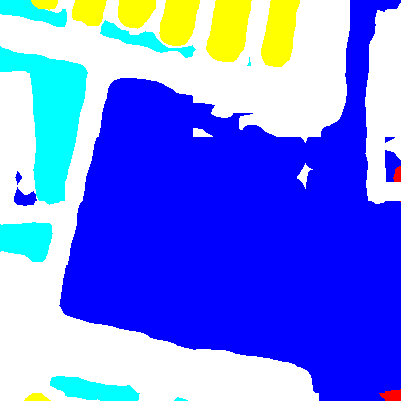}}
\vspace{-0.8em}

\subfloat{\includegraphics[width=0.112\textwidth]{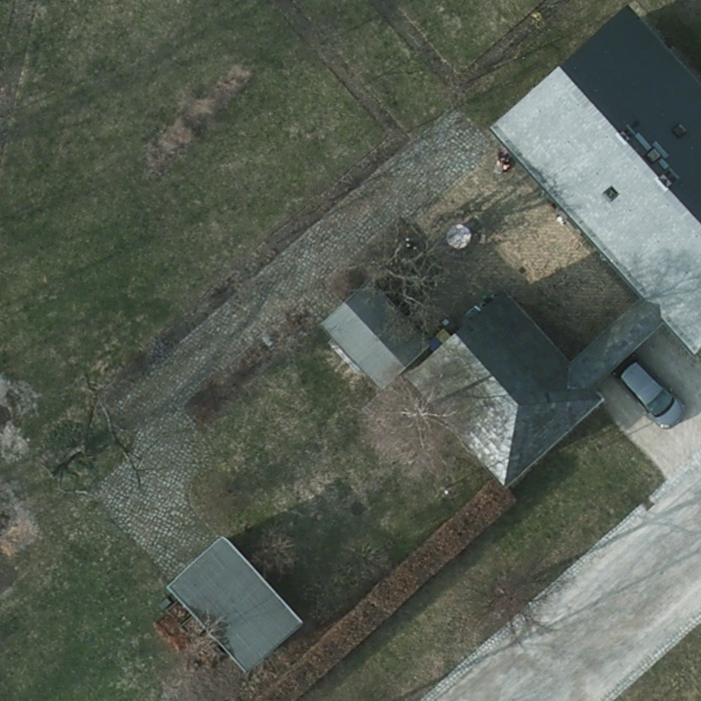}}
\hspace{-0.1em}
\subfloat{\includegraphics[width=0.112\textwidth]{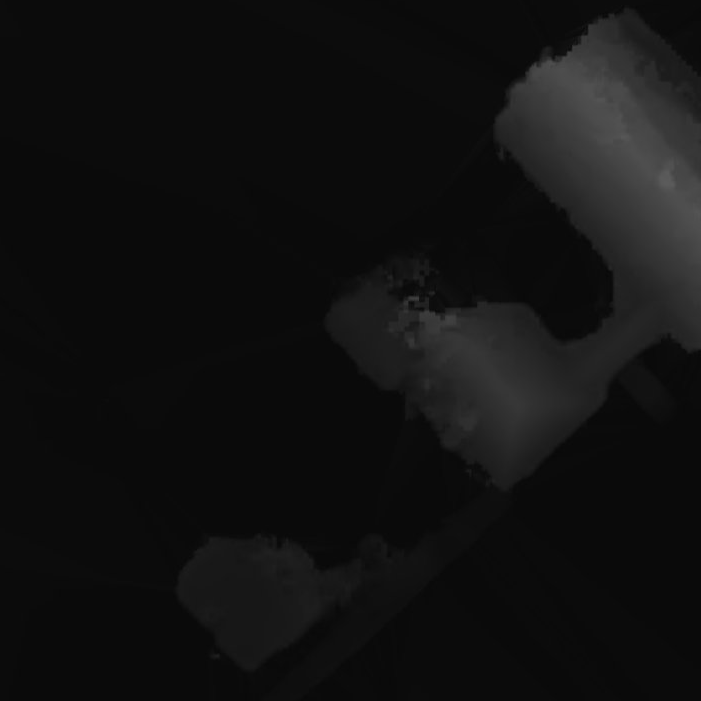}}
\hspace{-0.1em}
\subfloat{\includegraphics[width=0.112\textwidth]{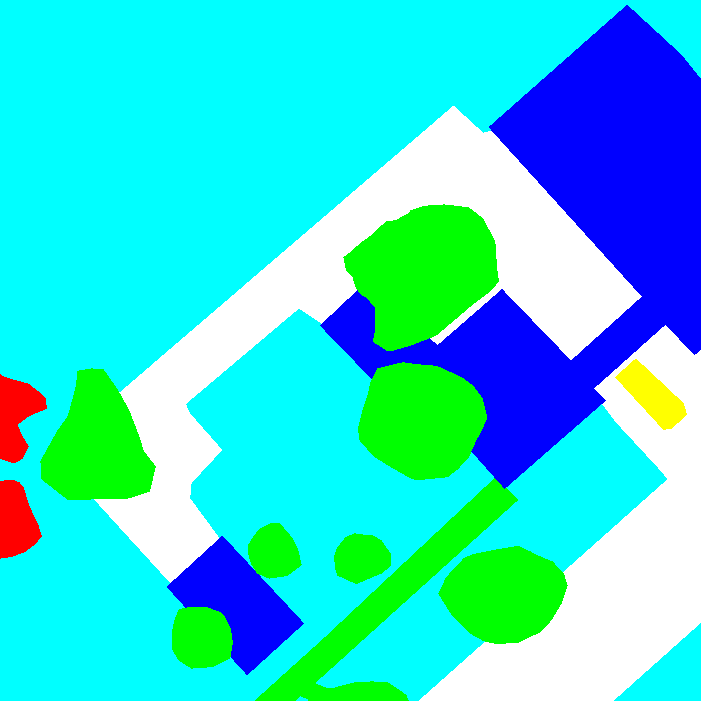}}
\hspace{-0.1em}
\subfloat{\includegraphics[width=0.112\textwidth]{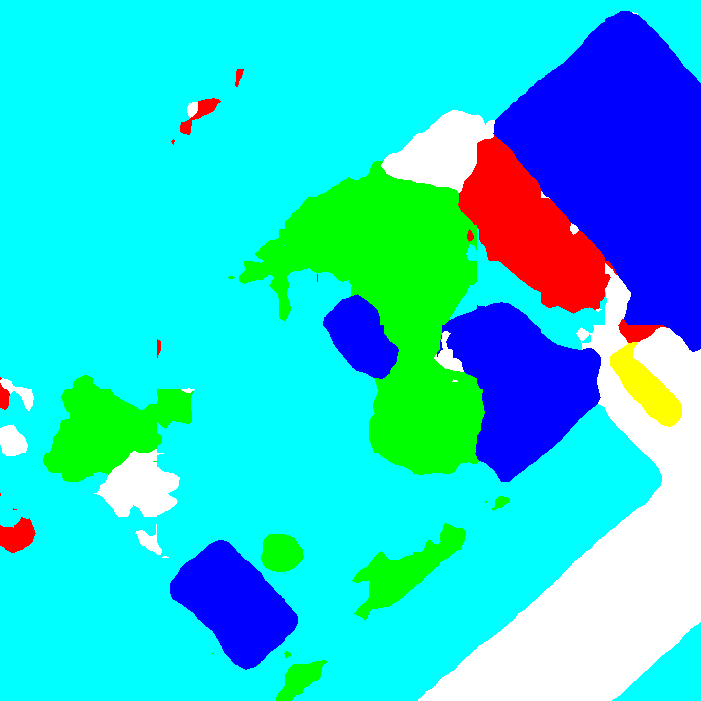}}
\hspace{-0.1em}
\subfloat{\includegraphics[width=0.112\textwidth]{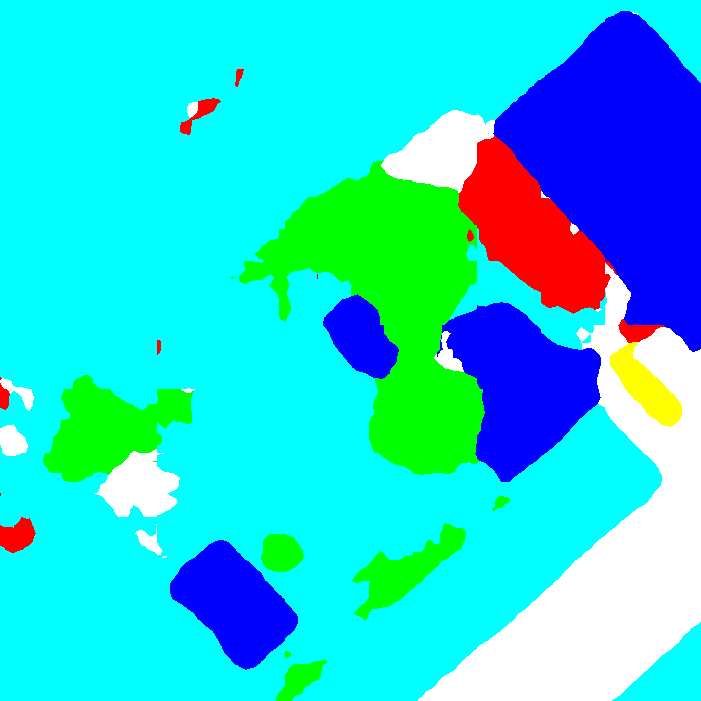}}
\hspace{-0.1em}
\subfloat{\includegraphics[width=0.112\textwidth]{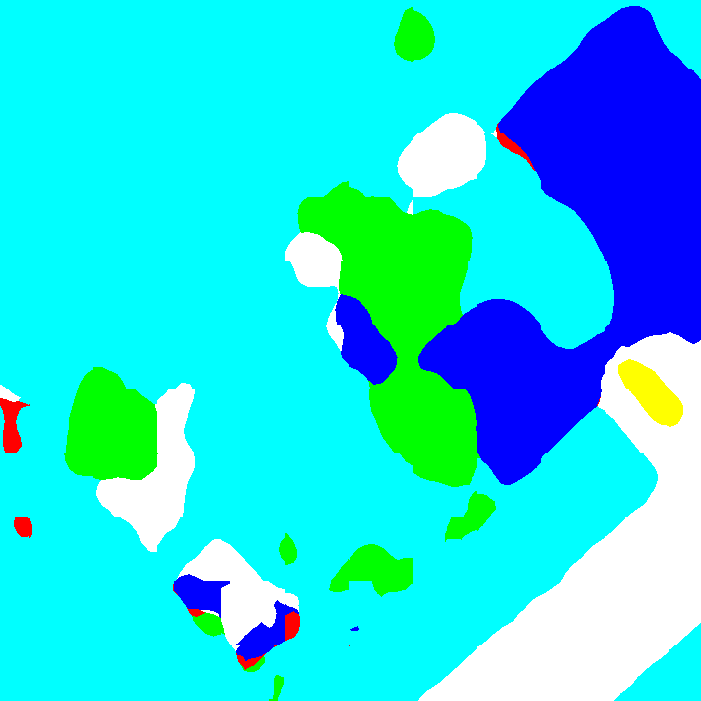}}
\hspace{-0.1em}
\subfloat{\includegraphics[width=0.112\textwidth]{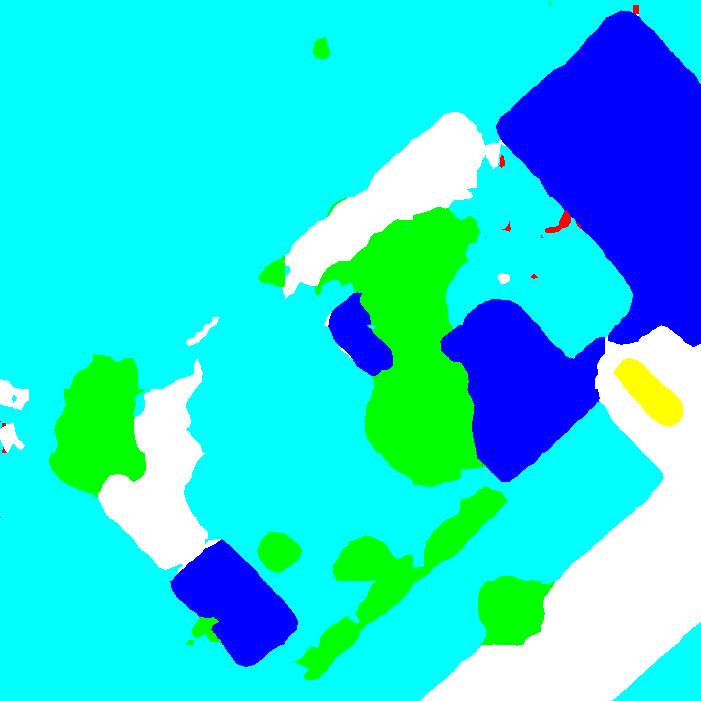}}
\hspace{-0.1em}
\subfloat{\includegraphics[width=0.112\textwidth]{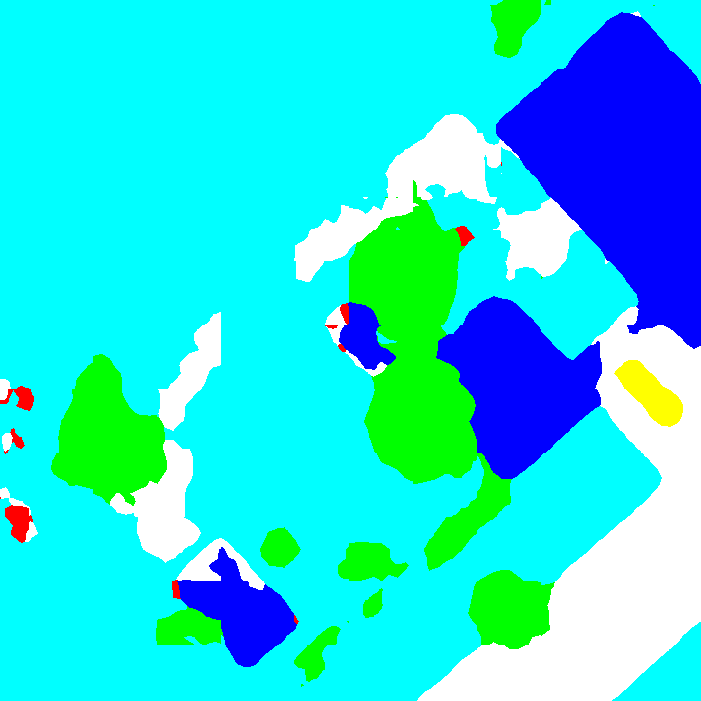}}
\vspace{-0.8em}

\subfloat[Image]{\includegraphics[width=0.112\textwidth]{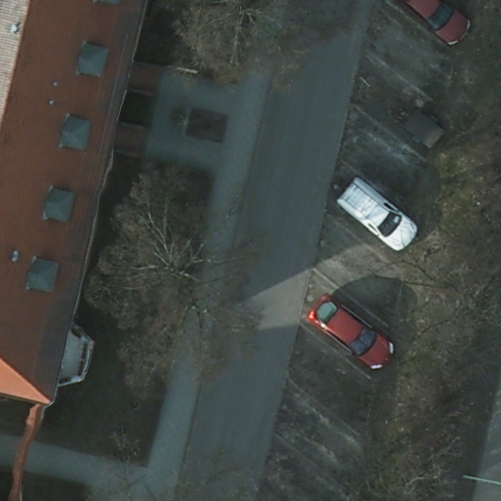}}
\hspace{-0.1em}
\subfloat[nDSM]{\includegraphics[width=0.112\textwidth]{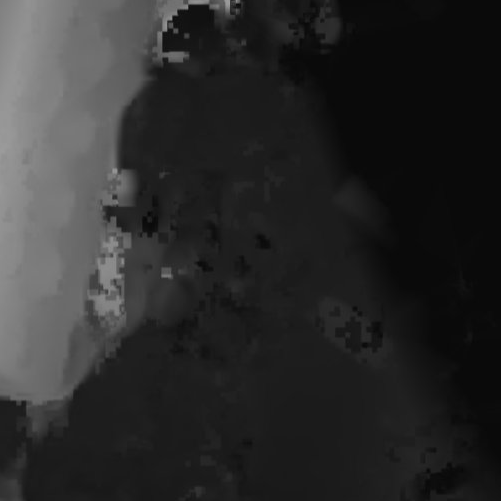}}
\hspace{-0.1em}
\subfloat[Ground Truth]{\includegraphics[width=0.112\textwidth]{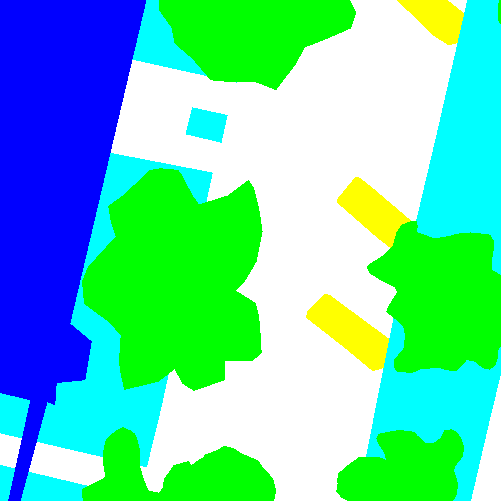}}
\hspace{-0.1em}
\subfloat[FCN]{\includegraphics[width=0.112\textwidth]{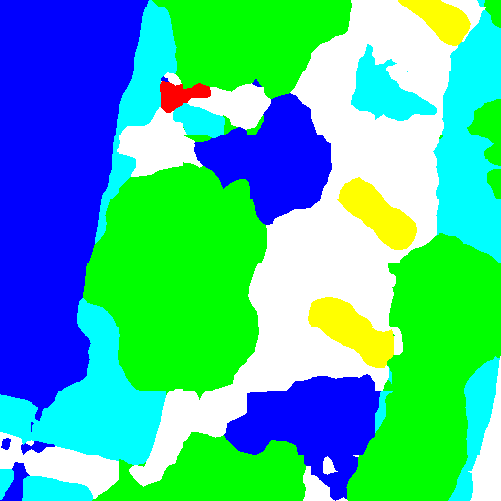}}
\hspace{-0.1em}
\subfloat[FCN-dCRF]{\includegraphics[width=0.112\textwidth]{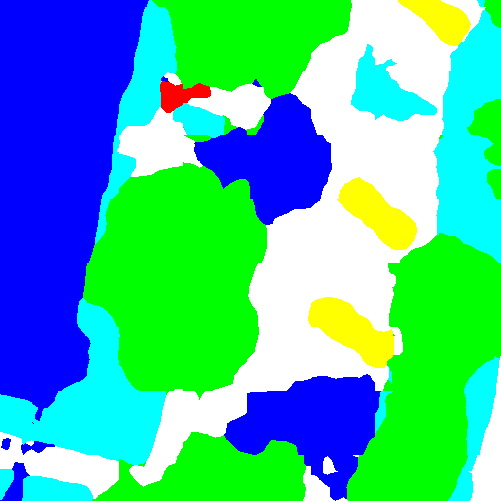}}
\hspace{-0.1em}
\subfloat[SCNN]{\includegraphics[width=0.112\textwidth]{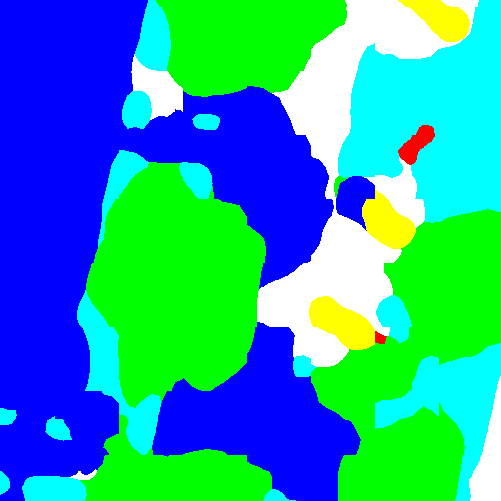}}
\hspace{-0.1em}
\subfloat[RA-FCN-srm]{\includegraphics[width=0.112\textwidth]{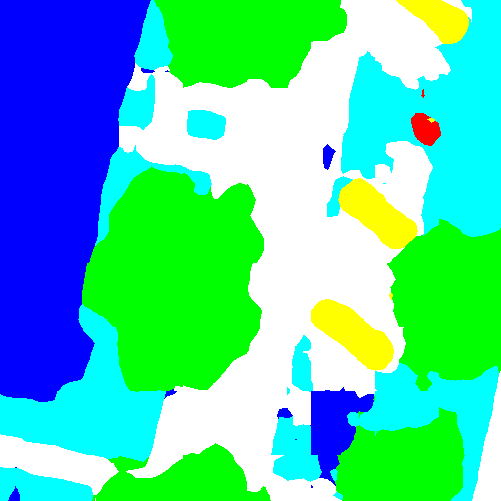}}
\hspace{-0.1em}
\subfloat[RA-FCN]{\includegraphics[width=0.112\textwidth]{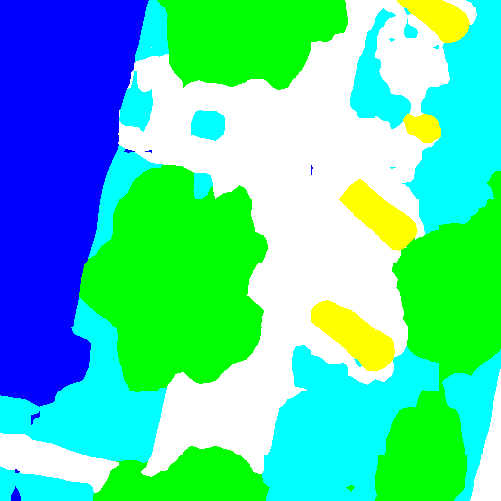}}

\caption{Examples of segmentation results on the Potsdam dataset. Legend---white: impervious surfaces, \textcolor{blue}{blue}: buildings, \textcolor{cyan}{cyan}: low vegetation, \textcolor{green}{green}: trees, \textcolor{yellow}{yellow}: cars, \textcolor{red}{red}: clutter/background.}
\label{fig:visual2}
\end{figure*}

Moreover, by taking advantage of spatial relation-augmented and channel relation-augmented features simultaneously, the performance of our network is further boosted up. The parallel integration of relation modules brings increments of 1.26\% and 0.14\% in the mean $F_1$ score with respect to RA-FCN-crm and RA-FCN-srm. Besides, a serial aggregation strategy is discussed, and results demonstrate that it behaves superiorly as compared to other models. To be more specific, such design achieves the highest mean $F_1$ score, 88.54\%, as well as the highest overall accuracy, 89.23\%. To conclude, spatial- and channel-augmented features extracted from relation modules carry out not only high-level semantics but also global relations in spatial and channel dimensionalities, which reinforces the performance of a network for semantic segmentation in aerial scenes.
\par

\subsection{Comparing with Existing Works}
For a comprehensive evaluation, we compare our model with six existing methods, including FCN~\cite{long2015fully}, FCN with fully connected CRF (FCN-dCRF)~\cite{deeplab}, spatial propagation CNN (SCNN)~\cite{pan2017spatial}, FCN with atrous convolution (Dilated FCN)~\cite{deeplab}, FCN with feature rearrangement (FCN-FR)~\cite{Maggiori2017HighResolutionAI}, CNN with full patch labeling by learned
upsampling (CNN-FPL)~\cite{volpi2017dense}, RotEqNet~\cite{marcos2018land}, PSPNet with VGG16 as backbone~\cite{cas_l4}, and several traditional methods~\cite{svl_5, hust}.

\begin{figure*}[!t]
\captionsetup[subfigure]{labelformat=empty, captionskip=2pt}
\centering
\subfloat[Image]{\includegraphics[width=0.23\textwidth]{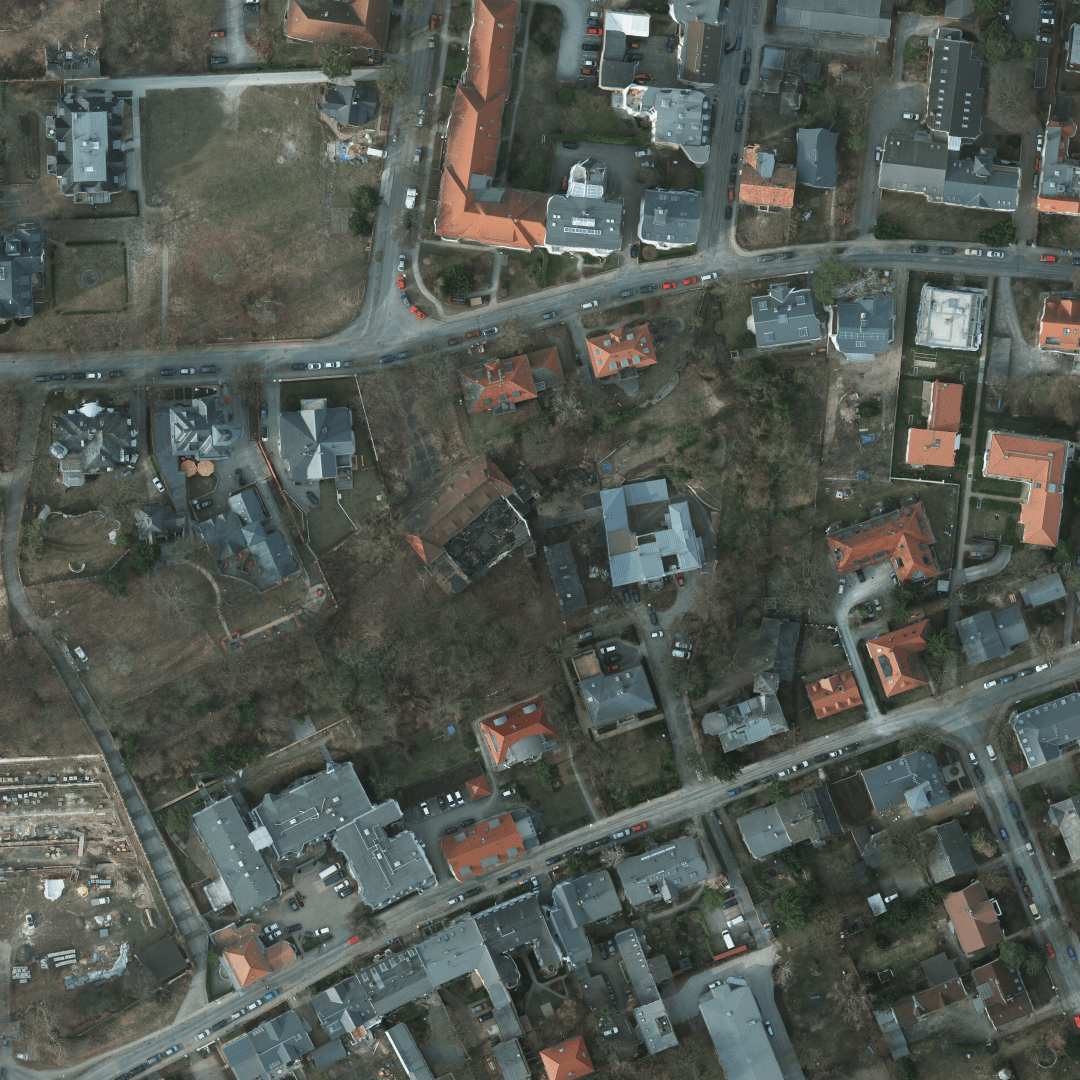}}
\hspace{0.1em}
\subfloat[Ground truth]{\includegraphics[width=0.23\textwidth]{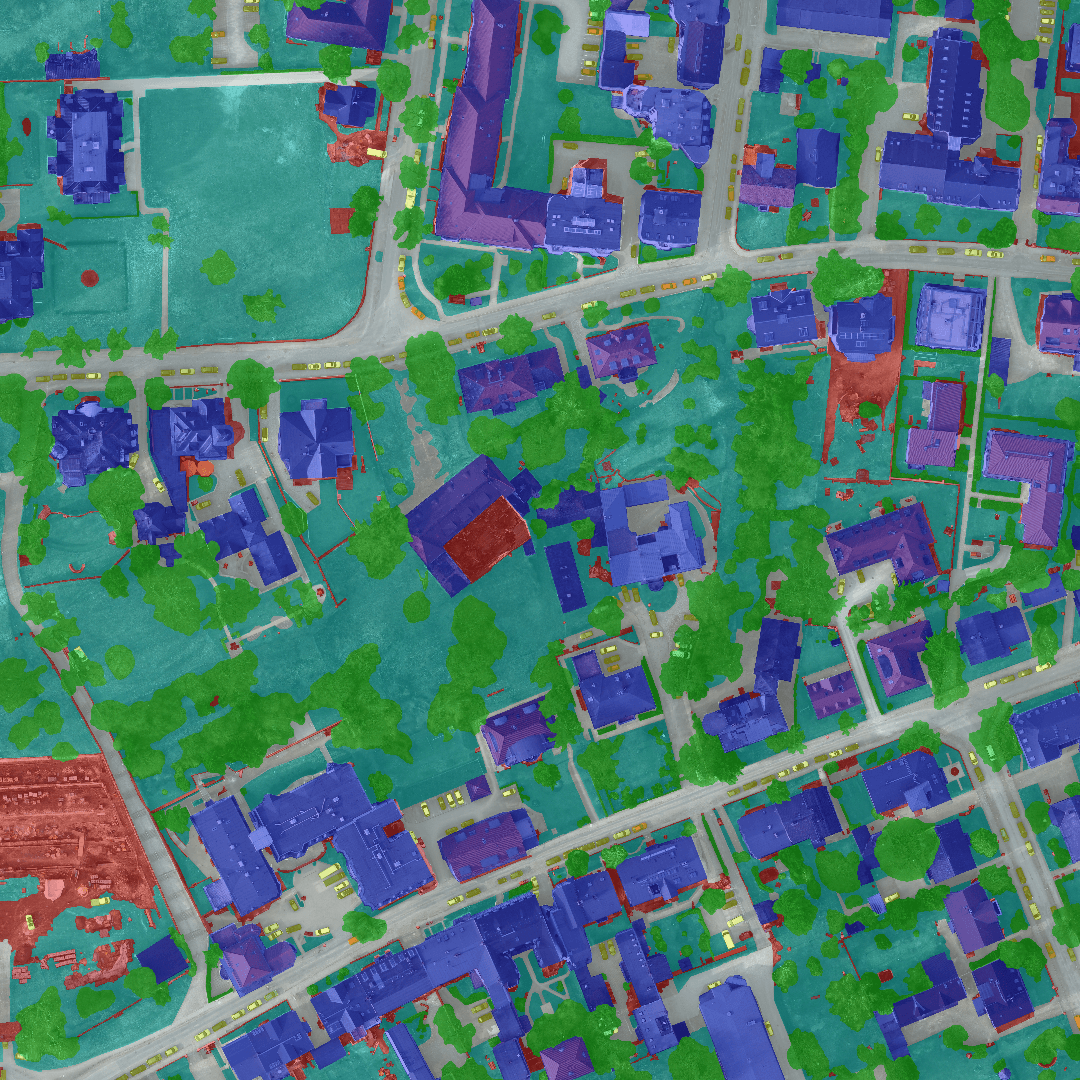}}
\hspace{0.1em}
\subfloat[FCN]{\includegraphics[width=0.23\textwidth]{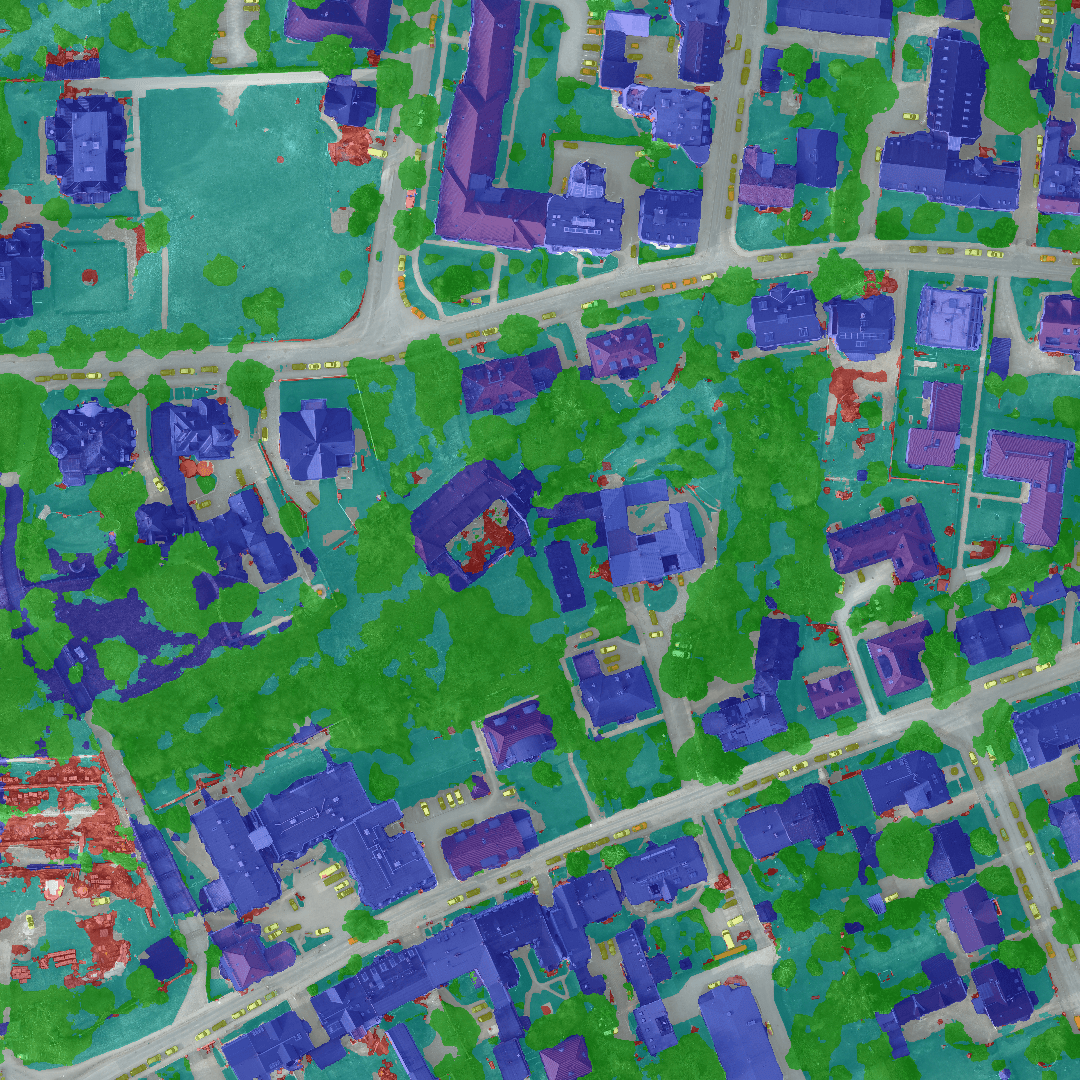}}
\vspace{-0.8em}

\subfloat[FCN-dCRF]{\includegraphics[width=0.23\textwidth]{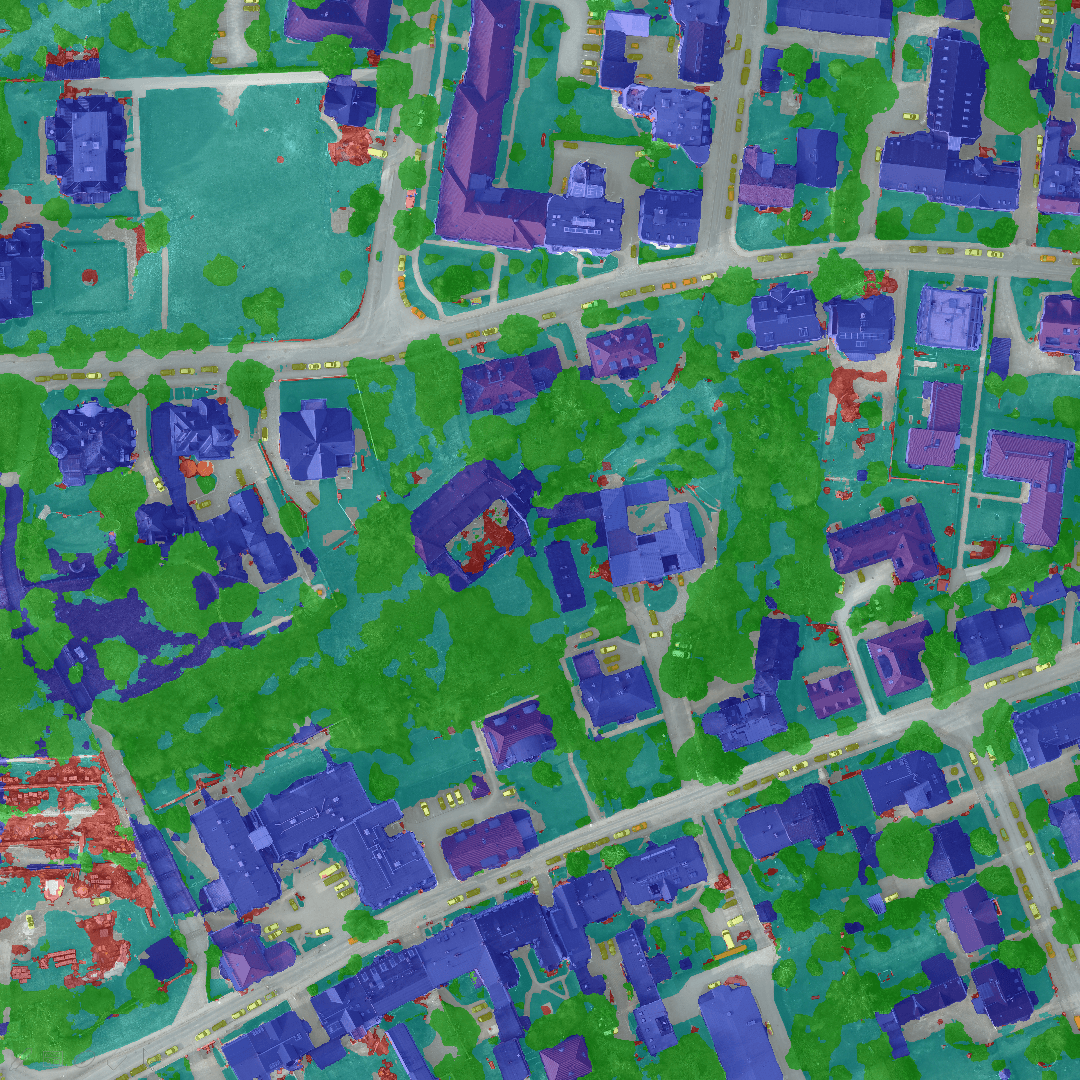}}
\hspace{0.1em}
\subfloat[SCNN]{\includegraphics[width=0.23\textwidth]{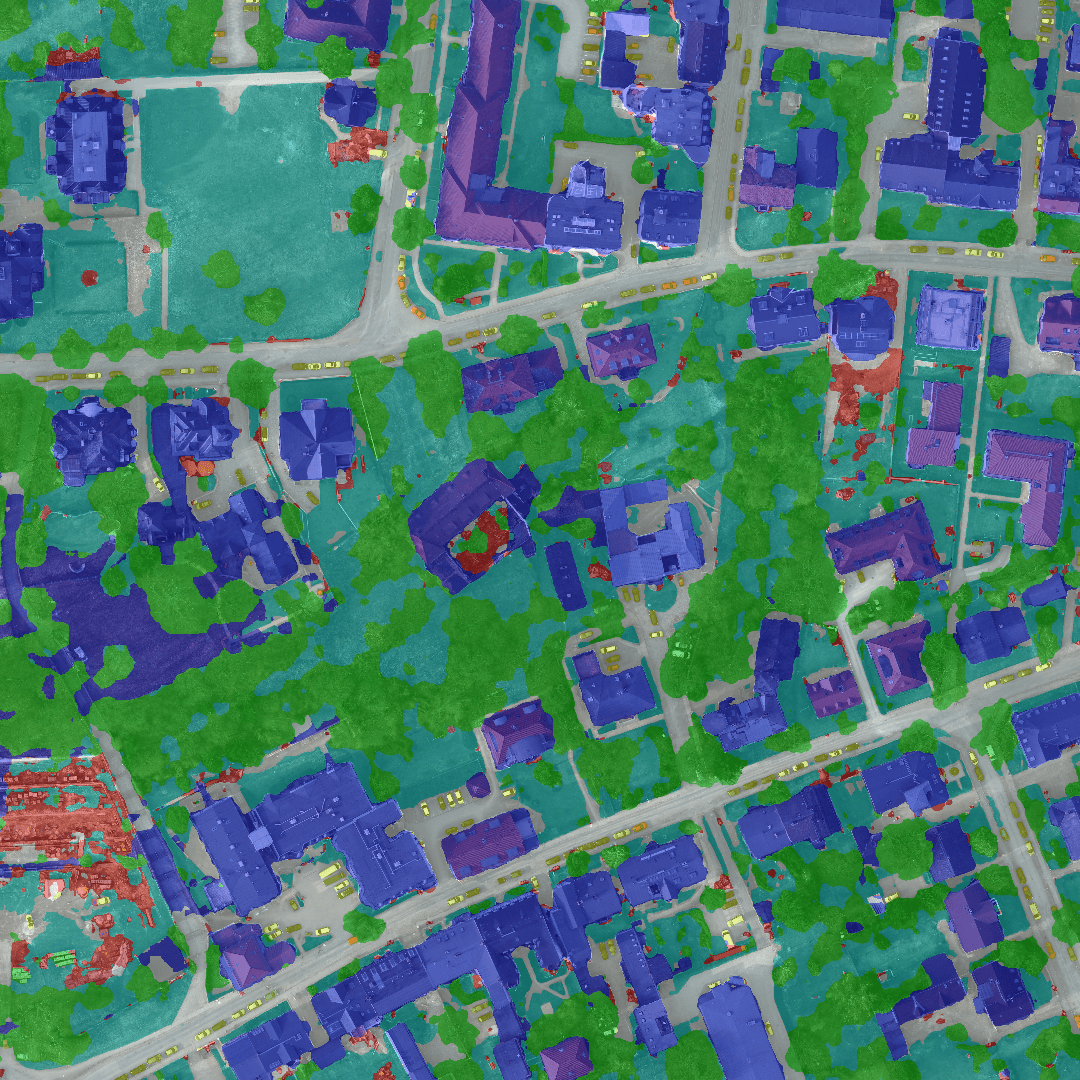}}
\hspace{0.1em}
\subfloat[RA-FCN]{\includegraphics[width=0.23\textwidth]{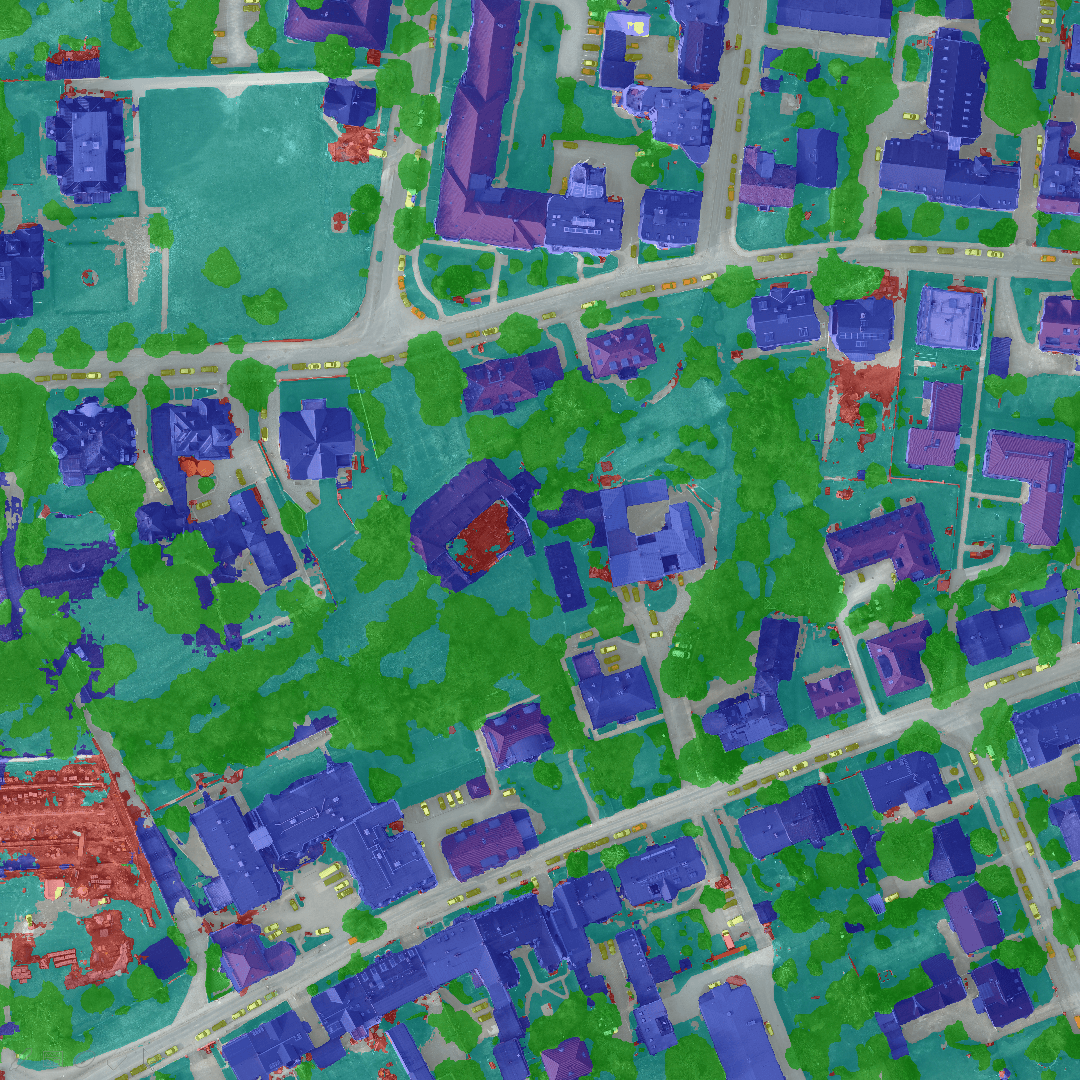}}
\renewcommand{\figurename}{Fig}
\caption{Example segmentation results of an image in the test set on Potsdam dataset ($90,000$ m$^2$). Legend---white: impervious surfaces, \textcolor{blue}{blue}: buildings, \textcolor{cyan}{cyan}: low vegetation, \textcolor{green}{green}: trees, \textcolor{yellow}{yellow}: cars, \textcolor{red}{red}: clutter/background. Zoom in for details.}
\label{fig:big}
\end{figure*}

Numerical results on the Vaihingen dataset are shown in Table~\ref{tab:vaihingen}. It is demonstrated that RA-FCN outperforms other methods in terms of mean $F_1$ score, mean IoU, and overall accuracy. Specifically, comparisons with FCN-dCRF and SCNN, where RA-FCN-srm obtains increments of 4.98\% and 3.69\% in mean $F_1$ score, respectively, validate the high performance of the spatial relation module in our network. Besides, compared to FCN-FR, RA-FCN reaches improvements of 1.96\% and 1.57\% in mean $F_1$ score and overall accuracy, which indicates the effectiveness of integrating the spatial relation module and channel relation module. Furthermore, per-class $F_1$ scores are calculated to assess the performance of recognizing different objects. It is noteworthy that our method remarkably surpasses other competitors in identifying scattered cars for its capacity of capturing long-range spatial relation.

\subsection{Qualitative Results}

Fig.~\ref{fig:visual1} shows a few examples of segmentation results. The second row demonstrates that networks with local receptive fields or relying on fully connected CRFs and spatial propagation modules fail to recognize impervious surfaces between two buildings, whereas our models make relatively accurate predictions. This is mainly because in this scene, the appearance of impervious surfaces is highly similar to that of the right building, which leads to a misjudgment of rival models. Thanks to the spatial relation module, RA-FCN-srm or RA-FCN is able to effectively capture useful visual cues from more remote regions in the image for an accurate inference. Besides, examples in the third row illustrate that RA-FCN is capable of identifying dispersively distributed objects as expected.

\subsection{Results on the Potsdam Dataset}
In order to further validate the effectiveness of our network, we conduct experiments on the Potsdam dataset, and numerical results are shown in Table~\ref{tab:potsdam}. The spatial relation module contributes to improvements of 2.25\% and 2.67\% in the mean $F_1$ score with respect to FCN-dCRF and SCNN, and the serial integration of both relation modules brings increments of 1.39\% and 1.54\% in the mean $F_1$ score, mean IoU, and overall accuracy, respectively.
\par
Moreover, qualitative results are presented in Figure~\ref{fig:visual2}. As shown in the first row, although low vegetation regions comprise intricate local contextual information and are liable to be misidentified, RA-FCN obtains more accurate results in comparison with other methods due to its remarkable capacity of exploiting global relations to solve visual ambiguities. The fourth row illustrates that outliers, i.e., the misclassified part of the building, can be eliminated by RA-FCN, while it is not easy for other competitors. To provide a thorough view of the performance of our network, we also exhibit a large-scale aerial scene as well as semantic segmentation results in Figure~\ref{fig:big}.

\section{Conclusion}
\label{sec:conc}
In this paper, we have introduced two effective network modules, namely the spatial relation module and the channel relation module, to enable relational reasoning in networks for semantic segmentation in aerial scenes. The comprehensive ablation experiments on aerial datasets where long-range spatial relations exist suggest that both relation modules have learned global relation information between objects and feature maps. However, our understanding of how these relation modules work for segmentation problems is preliminary and left as future works.
{\small
\bibliographystyle{ieee_fullname}
\bibliography{egbib}

\begin{thebibliography}{10}\itemsep=-1pt

\bibitem{7803544}
V. Badrinarayanan, A. Kendall, and R. Cipolla.
\newblock Segnet: A deep convolutional encoder-decoder architecture for image
  segmentation.
\newblock {\em IEEE Transactions on Pattern Analysis and Machine Intelligence},
  39(12):2481--2495, 2017.

\bibitem{Bell16}
S. Bell, C.~Lawrence Zitnick, K. Bala, and R. Girshick.
\newblock Inside-outside net: {Detecting} objects in context with skip pooling
  and recurrent neural networks.
\newblock In {\em {IEEE} International Conference on Computer Vision and
  Pattern Recognition (CVPR)}, 2016.

\bibitem{Bertasius16}
G. Bertasius, J. Shi, and L. Torresani.
\newblock Semantic segmentation with boundary neural fields.
\newblock In {\em {IEEE} International Conference on Computer Vision and
  Pattern Recognition (CVPR)}, 2016.

\bibitem{Chen16}
L.-C. Chen, J.~T. Barron, G. Papandreou, K. Murphy, and A.~L. Yuille.
\newblock Semantic image segmentation with task-specific edge detection using
  {CNNs} and a discriminatively trained domain transform.
\newblock In {\em {IEEE} International Conference on Computer Vision and
  Pattern Recognition (CVPR)}, 2016.

\bibitem{deeplab}
L.-C. Chen, G. Papandreou, I. Kokkinos, K. Murphy, and A.~L. Yuille.
\newblock {DeepLab}: {Semantic} image segmentation with deep convolutional
  nets, atrous convolution, and fully connected {CRFs}.
\newblock {\em arXiv:1606.00915}, 2016.

\bibitem{Chengeccv18}
X. Cheng, P. Wang, and R. Yang.
\newblock Depth estimation via affinity learned with convolutional spatial
  propagation network.
\newblock In {\em European Conference on Computer Vision (ECCV)}, 2018.

\bibitem{imagenet_cvpr09}
J. Deng, W. Dong, R. Socher, L.-J. Li, K. Li, and L. Fei-Fei.
\newblock Imagenet: A large-scale hierarchical image database.
\newblock In {\em {IEEE} International Conference on Computer Vision and
  Pattern Recognition (CVPR)}, 2009.

\bibitem{nadam2}
T. Dozat.
\newblock Incorporating {Nesterov} momentum into {Adam}.
\newblock 2015.

\bibitem{koller}
N. Friedman and D. Koller.
\newblock Being {B}ayesian about network structure. a {B}ayesian approach to
  structure discovery in {B}ayesian networks.
\newblock {\em Machine Learning}, 50(1-2):95--125, 2003.

\bibitem{svl_5}
M. Gerke.
\newblock {\em Use of the {Stair Vision Library} within the {ISPRS} {2D}
  Semantic Labeling Benchmark ({V}aihingen)}.
\newblock 2015.

\bibitem{Gould}
S. Gould, J. Rodgers, D. Cohen, G. Elidan, and D. Koller.
\newblock Multi-class segmentation with relative location prior.
\newblock {\em International Journal of Computer Vision}, 80(3):300--316, 2008.

\bibitem{he2016deep}
K. He, X. Zhang, S. Ren, and J. Sun.
\newblock Deep residual learning for image recognition.
\newblock In {\em {IEEE} International Conference on Computer Vision and
  Pattern Recognition (CVPR)}, 2016.

\bibitem{relationdet}
H. Hu, J. Gu, Z. Zhang, J. Dai, and Y. Wei.
\newblock Relation networks for object detection.
\newblock In {\em {IEEE} International Conference on Computer Vision and
  Pattern Recognition (CVPR)}, 2018.

\bibitem{SENet}
J. Hu, L. Shen, and G. Sun.
\newblock Squeeze-and-excitation networks.
\newblock In {\em {IEEE} International Conference on Computer Vision and
  Pattern Recognition (CVPR)}, 2018.

\bibitem{hua2019recurrently}
Y. Hua, L. Mou, and X.~X. Zhu.
\newblock Recurrently exploring class-wise attention in a hybrid convolutional
  and bidirectional {LSTM} network for multi-label aerial image classification.
\newblock {\em ISPRS Journal of Photogrammetry and Remote Sensing},
  149:188--199, 2019.

\bibitem{huang2017densely}
G. Huang, Z. Liu, L.~Van~Der Maaten, and K.~Q. Weinberger.
\newblock Densely connected convolutional networks.
\newblock In {\em {IEEE} International Conference on Computer Vision and
  Pattern Recognition (CVPR)}, 2017.

\bibitem{Kaiser17}
P. Kaiser, J.~D. Wegner, A. Lucchi, M. Jaggi, T. Hofmann, and K. Schindler.
\newblock Learning aerial image segmentation from online maps.
\newblock {\em IEEE Transactions on Geoscience and Remote Sensing},
  55(11):6054--6068, 2017.

\bibitem{Keeccv18}
T.-W. Ke, J.-J. Hwang, Z. Liu, and S.~X. Yu.
\newblock Adaptive affinity fields for semantic segmentation.
\newblock In {\em European Conference on Computer Vision (ECCV)}, 2018.

\bibitem{krizhevsky2012imagenet}
A. Krizhevsky, I. Sutskever, and G.~E. Hinton.
\newblock Imagenet classification with deep convolutional neural networks.
\newblock In {\em Advances in Neural Information Processing Systems (NIPS)},
  2012.

\bibitem{Liunips17}
S. Liu, S.~De Mello, J. Gu, G. Zhong, M.-H. Yang, and J. Kautz.
\newblock Learning affinity via spatial propagation networks.
\newblock In {\em Advances in Neural Information Processing Systems (NIPS)},
  2017.

\bibitem{Liueccv18}
S. Liu, G. Zhong, S.~De Mello, J. Gu, V. Jampani, M.-H. Yang, and J. Kautz.
\newblock Switchable temporal propagation network.
\newblock In {\em European Conference on Computer Vision (ECCV)}, 2018.

\bibitem{deepparsingnet}
Z. Liu, X. Li, P. Luo, C.~C. Loy, and X. Tang.
\newblock Semantic image segmentation via deep parsing network.
\newblock In {\em {IEEE} International Conference on Computer Vision and
  Pattern Recognition (CVPR)}, 2015.

\bibitem{long2015fully}
J. Long, E. Shelhamer, and T. Darrell.
\newblock Fully convolutional networks for semantic segmentation.
\newblock In {\em {IEEE} International Conference on Computer Vision and
  Pattern Recognition (CVPR)}, 2015.

\bibitem{Maggiori2017HighResolutionAI}
E. Maggiori, Y. Tarabalka, G. Charpiat, and P. Alliez.
\newblock High-resolution aerial image labeling with convolutional neural
  networks.
\newblock {\em IEEE Transactions on Geoscience and Remote Sensing},
  55(12):7092--7103, 2017.

\bibitem{Maire16}
M. Maire, T. Narihira, and S.~X. Yu.
\newblock Affinity {CNN}: {Learning} pixel-centric pairwise relations for
  figure/ground embedding.
\newblock In {\em {IEEE} International Conference on Computer Vision and
  Pattern Recognition (CVPR)}, 2016.

\bibitem{Marcoscvpr18}
D. Marcos, D. Tuia, B. Kellenberger, L. Zhang, M. Bai, R. Liao, and R. Urtasun.
\newblock Learning deep structured active contours end-to-end.
\newblock In {\em {IEEE} International Conference on Computer Vision and
  Pattern Recognition (CVPR)}, 2018.

\bibitem{marcos2018land}
D. Marcos, M. Volpi, B. Kellenberger, and D. Tuia.
\newblock Land cover mapping at very high resolution with rotation equivariant
  {CNN}s: Towards small yet accurate models.
\newblock {\em ISPRS Journal of Photogrammetry and Remote Sensing},
  145:96--107, 2018.

\bibitem{Marmanis18}
D. Marmanis, K. Schindler, J.~D. Wegner, S. Galliani, M. Datcu, and U. Stilla.
\newblock Classification with an edge: Improving semantic image segmentation
  with boundary detection.
\newblock {\em ISPRS Journal of Photogrammetry and Remote Sensing},
  135:158--172, 2018.

\bibitem{pan2017spatial}
X. Pan, J. Shi, P. Luo, X. Wang, and X. Tang.
\newblock Spatial as deep: {Spatial} {CNN} for traffic scene understanding.
\newblock In {\em {AAAI} Conference on Artificial Intelligence (AAAI)}, 2018.

\bibitem{hust}
N. Quang, N. Thuy, D. Sang, and H. Binh.
\newblock An efficient framework for pixel-wise building segmentation from
  aerial images.
\newblock In {\em International Symposium on Information and Communication
  Technology, ACM}, 2015.

\bibitem{santoro2017simple}
A. Santoro, D. Raposo, D.~G.T. Barrett, M. Malinowski, R. Pascanu, P.
  Battaglia, and T. Lillicrap.
\newblock A simple neural network module for relational reasoning.
\newblock In {\em Advances in Neural Information Processing Systems (NIPS)},
  2017.

\bibitem{sherrah2016fully}
J. Sherrah.
\newblock Fully convolutional networks for dense semantic labelling of
  high-resolution aerial imagery.
\newblock {\em arXiv:1606.02585}, 2016.

\bibitem{simonyan2014very}
K. Simonyan and A. Zisserman.
\newblock Very deep convolutional networks for large-scale image recognition.
\newblock {\em arXiv:1409.1556}, 2014.

\bibitem{vgg}
K. Simonyan and A. Zisserman.
\newblock Very deep convolutional networks for large-scale image recognition.
\newblock In {\em {IEEE} International Conference on Learning Representation
  (ICLR)}, 2015.

\bibitem{Tokarczyk15}
P. Tokarczyk, J.~D. Wegner, S. Walk, and K. Schindler.
\newblock Features, color spaces, and boosting: {New} insights on semantic
  classification of remote sensing images.
\newblock {\em IEEE Transactions on Geoscience and Remote Sensing},
  53(1):280--295, 2015.

\bibitem{volpi2017dense}
M. Volpi and D. Tuia.
\newblock Dense semantic labeling of subdecimeter resolution images with
  convolutional neural networks.
\newblock {\em IEEE Transactions on Geoscience and Remote Sensing},
  55(2):881--893, 2017.

\bibitem{Wainwright}
M.~J. Wainwright and M.~I. Jordan.
\newblock Graphical models, exponential families, and variational inference.
\newblock {\em Foundations and Trends in Machine Learning}, 1(1-2):1--305,
  2008.

\bibitem{torontocity}
S. Wang, M. Bai, G. Mattyus, H. Chen, W. Luo, B. Yang, J. Liang, J. Cheverie,
  S. Fidler, and R. Urtasun.
\newblock {TorontoCity}: Seeing the world with a million eyes.
\newblock In {\em {IEEE} International Conference on Computer Vision (ICCV)},
  2017.

\bibitem{fidler}
J. Yao, S. Fidler, and R. Urtasun.
\newblock Describing the scene as a whole: Joint object detection, scene
  classification and semantic segmentation.
\newblock In {\em CVPR}, 2012.

\bibitem{cas_l4}
H. Zhao, J. Shi, X. Qi, X. Wang, and J. Jia.
\newblock Pyramid scene parsing network.
\newblock In {\em {IEEE} International Conference on Computer Vision and
  Pattern Recognition (CVPR)}, 2017.

\bibitem{crfasrnn}
S. Zheng, S. Jayasumana, B. Romera-Paredes, V. Vineet, Z. Su, D. Du, C. Huang,
  and P.~H.~S. Torr.
\newblock Conditional random fields as recurrent neural networks.
\newblock In {\em {IEEE} International Conference on Computer Vision (ICCV)},
  2015.

\bibitem{zhou2017temporal}
B. Zhou, A. Andonian, and A. Torralba.
\newblock Temporal relational reasoning in videos.
\newblock In {\em European Conference on Computer Vision (ECCV)}, 2018.

\bibitem{peng2017large}
B. Zhou, A. Khosla, A. Lapedriza, A. Oliva, and A. Torralba.
\newblock Object detectors emerge in deep scene {CNNs}.
\newblock In {\em {IEEE} International Conference on Learning Representation
  (ICLR)}, 2015.

\bibitem{zhu2017deep}
X.~X. Zhu, D. Tuia, L. Mou, G. Xia, L. Zhang, F. Xu, and F. Fraundorfer.
\newblock Deep learning in remote sensing: A comprehensive review and list of
  resources.
\newblock {\em IEEE Geoscience and Remote Sensing Magazine}, 5(4):8--36, 2017.

\end{thebibliography}
}

\end{document}